\documentclass{ieeeaccess}
\usepackage{cite}
\usepackage{amsmath,amssymb,amsfonts}
\usepackage{algorithmic}
\usepackage{graphicx}
\usepackage{textcomp}
\usepackage[T1]{fontenc}
\usepackage{url}
\usepackage{pifont}
\usepackage{array}
\usepackage{multicol}
\usepackage{multirow}
\usepackage{subcaption}
\usepackage{makecell}
\usepackage{float}
\usepackage{adjustbox}
\usepackage{rotating}
\usepackage{booktabs}
\usepackage{balance}
\usepackage[table]{xcolor}
\usepackage{xcolor}
\usepackage{soul}
\usepackage{bm}
\makeatletter
\AtBeginDocument{\DeclareMathVersion{bold}
\SetSymbolFont{operators}{bold}{T1}{times}{b}{n}
\SetSymbolFont{NewLetters}{bold}{T1}{times}{b}{it}
\SetMathAlphabet{\mathrm}{bold}{T1}{times}{b}{n}
\SetMathAlphabet{\mathit}{bold}{T1}{times}{b}{it}
\SetMathAlphabet{\mathbf}{bold}{T1}{times}{b}{n}
\SetMathAlphabet{\mathtt}{bold}{OT1}{pcr}{b}{n}
\SetSymbolFont{symbols}{bold}{OMS}{cmsy}{b}{n}
\renewcommand\boldmath{\@nomath\boldmath\mathversion{bold}}}
\makeatother

\def\BibTeX{{\rm B\kern-.05em{\sc i\kern-.025em b}\kern-.08em
    T\kern-.1667em\lower.7ex\hbox{E}\kern-.125emX}}

\DeclareTextFontCommand{\textbf}{\bfseries}
\DeclareTextFontCommand{\textit}{\itshape}

\newcommand{\cmark}{\ding{51}}%
\newcommand{\xmark}{\ding{55}}

\newif\ifhighlight
\highlightfalse
\ifhighlight
\else
\renewcommand{\hl}[1]{#1}
\fi

\newcommand{\LMtraining}{\texttt{LMtr}}
\newcommand{\GANtraining}{\texttt{GANtr}}
\newcommand{\gpt}{GPT}
\newcommand{\transformer}{TRA}
\newcommand{\gru}{GRU}
\newcommand{\oneh}{\texttt{1h}}
\newcommand{\emb}{\texttt{emb}}
\newcommand{\transformerOnehot}{\transformer\textsubscript{\oneh}}
\newcommand{\gruOnehot}{\gru\textsubscript{\oneh}}
\newcommand{\transformerEmbed}{\transformer\textsubscript{\emb}}
\newcommand{\gruEmbed}{\gru\textsubscript{\emb}}
\newcommand{\gptEmbed}{\gpt\textsubscript{\emb}}
\newcommand{\svm}{SVM}
\newcommand{\svmLMgpt}{\svm+GPT$_{\emb}^{\LMtraining}$}
\newcommand{\svmGANgpt}{\svm+GPT$_{\emb}^{\GANtraining}$}
\newcommand{\svmLMonehotTrans}{\svm+\transformer$_{\oneh}^{\text{\LMtraining}}$}
\newcommand{\svmGANonehotTrans}{\svm+\transformer$_{\oneh}^{\text{\GANtraining}}$}
\newcommand{\svmLMonehotGru}{\svm+\gru$_{\oneh}^{\text{\LMtraining}}$}
\newcommand{\svmGANonehotGru}{\svm+\gru$_{\oneh}^{\text{\GANtraining}}$}
\newcommand{\nn}{CNN}
\newcommand{\nngpt}{\nn}
\newcommand{\nnLMgpt}{\nngpt+GPT$_{\emb}^{\LMtraining}$}
\newcommand{\nnGANgpt}{\nngpt+GPT$_{\emb}^{\GANtraining}$}
\newcommand{\nnonehot}{\nn}
\newcommand{\nnLMonehotTrans}{\nnonehot+\transformer$_{\oneh}^{\text{\LMtraining}}$}
\newcommand{\nnGANonehotTrans}{\nnonehot+\transformer$_{\oneh}^{\text{\GANtraining}}$}
\newcommand{\nnLMonehotGru}{\nnonehot+\gru$_{\oneh}^{\LMtraining}$}
\newcommand{\nnGANonehotGru}{\nnonehot+\gru$_{\oneh}^{\text{\GANtraining}}$}
\newcommand{\nnembed}{\nn}
\newcommand{\nnLMembedTrans}{\nnembed+\transformer$_{\emb}^{\text{\LMtraining}}$}
\newcommand{\nnGANembedTrans}{\nnembed+\transformer$_{\emb}^{\text{\GANtraining}}$}
\newcommand{\nnLMembedGru}{\nnembed+\gru$_{\emb}^{\text{\LMtraining}}$}
\newcommand{\nnGANembedGru}{\nnembed+\gru$_{\emb}^{\text{\GANtraining}}$}

\newif\ifdraft
\draftfalse

\begin{document}
\history{Date of publication xxxx 00, 0000, date of current version xxxx 00, 0000.}
\doi{xxxx}

\title{Forging the Forger: An Attempt to Improve Authorship Verification via Data Augmentation}

\author{\uppercase{Silvia Corbara}\authorrefmark{1,2},
\uppercase{Alejandro Moreo}\authorrefmark{2}} 
\address[1]{Scuola Normale Superiore, 56126 Pisa, IT (e-mail: name.surname@sns.it)}
\address[2]{Istituto di Scienza e Tecnologie dell'Informazione, Consiglio Nazionale delle Ricerche, 56124 Pisa, IT (e-mail: name.surname@isti.cnr.it)}

\markboth
{Silvia Corbara and Alejandro Moreo: AV via Data Augmentation}
{Silvia Corbara and Alejandro Moreo: AV via Data Augmentation}

\corresp{Corresponding author: Silvia Corbara (e-mail: silvia.corbara@isti.cnr.it).}

\begin{abstract}
\noindent Authorship Verification (AV) is a text classification task concerned with inferring whether a candidate text has been written by one specific author ($A$) or by someone else ($\overline{A}$).
It has been shown that many AV systems are vulnerable to adversarial attacks, where a malicious author actively tries to fool the classifier by either concealing their writing style, or by imitating the style of another author. In this paper, we investigate the potential benefits of augmenting the classifier training set with (negative) synthetic examples. These synthetic examples are generated to imitate the style of $A$. We analyze the improvements in \hl{the classifier predictions} that this augmentation brings to bear in the task of AV in an adversarial setting.
In particular, we experiment with three different generator architectures (one based on Recurrent Neural Networks, another based on small-scale transformers, and another based on the popular GPT model) and with two training strategies (one inspired by standard Language Models, and another inspired by Wasserstein Generative Adversarial Networks). We evaluate our hypothesis on five datasets (three of which have been  specifically collected to represent an adversarial setting) and using two learning algorithms for the AV classifier (Support Vector Machines and Convolutional Neural Networks). This experimentation \hl{yields} negative results, revealing that, although our methodology proves effective in many adversarial settings, its benefits are too sporadic for a pragmatical application.
\end{abstract}

\begin{keywords}
Authorship Identification, Authorship Verification, Data augmentation, Text classification
\end{keywords}

\titlepgskip=-21pt

\maketitle

\section{Introduction}
\label{sec:intro}
\noindent The field of \textit{Authorship Identification} (AId) is the branch of \textit{Authorship Analysis} concerned with the study of the true identity of the author of a written document of unknown or debated paternity. \textit{Authorship Verification} (AV) is one of the main tasks of AId: given a single candidate author $A$ and a document $d$, the goal is to infer whether $A$ is the real author of $d$ or not \cite{Stamatatos:2016ij}. The author $A$ is conventionally represented by a set of documents that we \hl{unequivocally} know have been written by $A$. AV \hl{is} sometimes cast as a one-class classification problem \cite{Stein:2008ma, Koppel:2007sa}, with $A$ as the only class. Nevertheless, it is more commonly addressed as a binary classification problem, with $A$ and $\overline{A}$ as the possible classes, where $\overline{A}$ is characterized by a collection of documents from authors other than $A$, but somehow related to $A$ (\hl{sharing, e.g.,} the same language, period, literary style, genre). 

The goal of AId in general, and of AV in particular, is to find a proper way to profile the ``hand'' of a given writer, \hl{in order} to clearly \hl{distinguish} their written production from that of other authors. This characterization is often \hl{tackled} through the use of ``stylometry'', a methodology that disregards the artistic value and meaning of a written work, in favour of conducting a frequentist analysis of linguistic events. These events, also known as ``style markers'', typically escape the conscious control of \hl{the} writer and are assumed to remain approximately invariant throughout the literary production of a given author, while conversely varying substantially across different authors \cite[p.\ 241]{Juola:2006jn}. Current approaches to AV \hl{typically} rely on \emph{automated text classification}, wherein a supervised machine learning algorithm is trained to generate a classifier that distinguishes the works of $A$ from those of other authors by \hl{analyzing the} vectorial representations of the documents that \hl{incorporate} different style-markers.

However, training and employing an effective classifier can be very \hl{challenging}, or even impossible, if an ``adversary'' is at play, i.e., when a human or an automatic process actively tries to mislead the classification. We can distinguish between two main variants of this adversarial setting \cite{Brennan:2012la}. In the first case, called ``obfuscation'', the authors themselves try to conceal their writing style in order to not be recognized. This can be done manually, but nowadays \hl{specific automatic tools are available for this purpose}; for example, by applying sequential steps of machine translation \cite{Faust:2017al}. In the second case, called ``imitation'', a forger (the ``imitator'') tries to replicate the writing style of another specific author. While the former case can be considered as possibly less harmful (a writer might simply be interested in preserving their privacy, without necessarily harbouring any malevolent intent), the latter case is inherently illicit (an exception to this may be found in artistic tributes --- provided they are accompanied by a statement openly acknowledging the intent).
Our cultural heritage is indeed filled with countless examples of historical documents of questioned authorship, often caused by supposed forgeries or false appropriations \cite{Corbara:2019cq, McCarthy:2021wt, Nini:2018pu, Savoy:2019pi, Tuccinardi:2017pp, Vainio:2019gr}. Moreover, due to the recent significant advances in language modelling, powerful Neural Networks (NNs) are now able to autonomously generate ``coherent, non-trivial and human-like'' text samples \cite[p.\ 1]{Fagni:2021tf}, that can be exploited as fake news or propaganda \cite{Fagni:2021tf, Salminen:2022ct}. Indeed, it has been shown that many AId systems can be easily deceived in adversarial contexts \cite{Brennan:2012la, Potthast:2016ha}. \hl{This has given rise to the discipline known as ``adversarial authorship'' (or ``adversarial stylometry'', or ``authorship obfuscation'') devoted to study specific techniques able to fool AId systems \mbox{{\cite{Bevendorff:2019ch, Faust:2017al, Allred:2020ts, Zhai:2022ig, Uchendu:2023tr, Wang:2023gf}}}.}

A seemingly straightforward solution to this problem \hl{would be to add a set of representative texts from the forger to} the training set used \hl{to build} the classifier, thus allowing the training algorithm \hl{to} find descriptive patterns that effectively identify adversarial examples. Unfortunately, this strategy is generally unfeasible, as in most cases we might not expect to have any such examples.%, or at best, have only one (i.e., in the seldom case where it has been proven that a document is from a forger). 
%This limitation makes it entirely impossible to discover statistically sound patterns when there are no positive examples, and nearly impossible at best when we have only one example. 

In this work, we investigate ways \hl{to improve} the performance of an AV classifier by augmenting its training set with \textit{synthetically} generated examples that mimic the style of the author that the classifier tries to identify. In particular, we explore various generator architectures, including a GRU model \cite{Cho:2014ps}, a standard transformer \cite{Vaswani:2017tt}, and the popular GPT system (in its shallower variant, \texttt{DistilGPT2} \cite{Radford:2019uo}). Two distinct training strategies for the generators are employed: one draws inspiration from standard Language Models (LMs) where the generator learns to emulate the writing style of a specific author, while the other takes cues from Wasserstein Generative Adversarial Networks (WGANs) by training a generator to exploit the weaknesses of the classifier. We run experiments on five datasets (three of them collected specially to simulate an adversarial setting), and using two learning algorithms for the AV classifier: Support Vector Machines (SVMs) and Convolutional Neural Networks (CNNs).

The remaining of this paper is organized as follows. In Section~\ref{sec:related_work} we survey the related literature. In Section~\ref{sec:method}, we present our method, describing the generator architectures and their training strategies, and the learning algorithms that we employ for the AV classifier. In Section~\ref{sec:exp}, we first detail the experimental setting, including the datasets and the experimental protocol we use, and then we present and comment on the results of our experiments. Despite our efforts, the results we \hl{obtain} seem to indicate that data augmentation is not always beneficial for the task of AV \hl{(at least with the techniques we study here)}; in Section~\ref{subsec:exp_expl} we analyse the possible causes of the negative results. Finally, Section~\ref{sec:conclusion} wraps up, offering some final remarks and pointing to some possible avenues for future research.

\section{Related work}
\label{sec:related_work}
\noindent AId is usually tackled by means of machine-learned classifiers or distance-based methods; the annual PAN shared tasks \cite{Kestemont:2019wo, Bevendorff:2020hn, Bevendorff:2021iw, Stamatatos:2022or, bevendorff2023overview} offer a very good overview of the most recent trends in the field.

\hl{In particular,} the baselines presented in the 2019 edition \cite{Kestemont:2019wo} are \hl{representative of the systems that have long been considered standard}, i.e., traditional learning algorithms such as SVM or logistic regression, compression-based algorithms, and variations on the well-known Impostors Method \cite{Koppel:2014di}. In particular, SVM has become a standard learning algorithm for many text classification tasks, due to its robustness to high dimensionality and to its wide scope of applicability. In various settings, SVMs have been found to outperform other learning algorithms such as decision trees and even NNs \cite{Zheng:2006wf}, especially in regimes of data scarcity \cite{Boran:2020hp}. On the other hand, despite the many successes achieved in other natural language processing tasks \cite{Young:2018tn}, deep NNs have rarely been employed in tasks of AId, arguably due to the huge quantity of training data they usually require. Even though one of the first appearances of NNs at PAN dates back to 2015 \cite{Bagnall:2015ro}, and even though this approach won the competition \cite{Stamatatos:2015nv}, \hl{only recently} the generalized belief that ``simple approaches based on character/word $n$-grams and well-known classification algorithms are much more effective in this task than more sophisticated methods based on deep learning'' \cite[p.9]{Kestemont:2018vo} \hl{was} called into question. As a result, NNs methods are nowadays becoming more and more \hl{common} at PAN \cite{Bevendorff:2020hn, Bevendorff:2021iw, Stamatatos:2022or}.

Needless to say, the use of data augmentation for improving classification performance is not a new idea, neither in the text classification field nor in the AId field. Researchers have explored various techniques for generating synthetic samples, such as the random combination of real texts \cite{Theophilo:2021oi, Boenninghoff:2019mi}, or the random substitutions of words with synonyms  \cite{Zhang:2015rc}, or by interrogation of LMs \cite{Kobayashi:2018tc}. Similarly, adversarial examples are generated by a process that actively tries to fool the classification \cite{Goodfellow:2014jr}, and have been extensively used to improve the training of a classifier for many text classification tasks in general, and for counteracting adversarial attacks in particular. For example, Zhai et al.~\cite{Zhai:2022ig} feed the learning algorithm with (real) texts that have been purposefully obfuscated, in order to make the classifier robust to obfuscation. 

Unlike these works, we do not automatically modify pre-existing texts; instead, we employ various generation algorithms to create new samples, simulating the deceitful actions of an adversary. In particular, one generative technique we explore is the so-called Generative Adversarial Network (GAN), which was first introduced by Goodfellow et al.~\cite{Goodfellow:2014gn} back in 2014. The GAN architecture is made of two components: a Generator ($G$) that produces synthetic (hence fake) examples, and a Discriminator ($D$) that classifies examples as ``real'' (i.e., coming from a real-world distribution) or ``fake'' (i.e., generated by $G$). Both components play a min-max game, where $D$ tries to correctly spot the examples created by $G$, while $G$ tries to produce more plausible examples in order to fool $D$. %More formally:

%\begin{equation}
%\min_G \max_D V(D, G) = \mathbb{E}_{x \sim p_{\text{data}}}[\log D(x)] + \mathbb{E}_{z \sim p_{\text{noise}}}[\log(1 - D(G(z)))]
%\end{equation}
%
%\noindent where $D(x)$ is the posterior probability $D$ assigns to the fact that $x$ is a real example, and $D(G(z))$ is the posterior probability that a fake instance (i.e., a datapoint created by the Generator) is real. 

Although the GAN strategy has excelled in the generation of images \cite{Karras:2019ty}, its application to text generation has proven rather cumbersome. One of the main reasons behind this is the discrete nature of language: choosing the next word in a sequence implies picking one symbol from a vocabulary, an operation that is typically carried out through an argmax operation, which blocks the gradient flow during backpropagation. Some strategies have been proposed to overcome this limitation. One idea is to employ a reinforcement learning approach, where the feedback from $D$ acts as reward \cite{Yu2017:qe}, although this strategy has been found to be inefficient to train  \cite{Donahue:2018iv}. An alternative idea is to use the Gumbel-Softmax operation to obtain a differentiable approximation of one-hot vectors \cite{Kusner:2016sg}, or to directly \hl{process} the continuous output of the generator (i.e., refraining from choosing specific tokens); some examples include the encoding of real sentences via an autoencoder \cite{Donahue:2018iv}, or the creation of a word-embedding matrix for real sentences \cite{Zhang:2017la}. 

The work by Hatua et al.~\cite{Hatua:2021tb} bears some similarities with our methodology: they employ a GAN-trained generator to produce new examples, which are labelled as negatives and then added to the classifier training dataset. Their experiments indeed \hl{demonstrate} the benefits of data augmentation for the fact-checking task.

\hl{Within the AId field, the work by Manjavacas et al.\mbox{~\cite{Manjavacas:2017as}} explores the idea to use data augmentation to boost the performance of an authorship classifier, employing a RNN as language model. They indeed report a slight increase in the attribution performance of the system; however, they perform a very narrow experimentation, without significance testing, and do not tackle the problem of forgery, limiting their investigation to a single non-adversarial setting.}

Finally, this paper is a thorough extension of the preliminary experiments presented in \hl{the} short paper by Corbara and Moreo~\cite{Corbara:2023gan}.
In the present work, we carry out a more robust experimentation, in which we test an improved GAN strategy (based on the Wasserstein GAN), and in which we consider different vector representations for the input of our generator models. We also expand the number of datasets from three to five.

\section{Methodology}
\label{sec:method}
\noindent In this paper, \hl{we approach the AV task as a binary text classification problem where the objective is to determine whether a given document $d$ was written by a specific candidate author (representing the positive class $A$, which includes representative texts from that author) or by any other author (collectively represented as the negative class $\overline{A}$, encompassing examples from other authors)}. The union of all \hl{the documents} with \hl{the} corresponding labels define the training set $L$ that we use \hl{to generate} a classifier $h$ via an inductive algorithm.

We propose to enhance the classifier performance with the addition of adversarial training examples, i.e., textual examples specifically generated to imitate the author $A$. The generated examples are labelled as negative examples ($\overline{A}$) and added to $L$ with the \hl{aim} of generating an improved classifier $h^*$.
Note that, unlike works such as the one by Jones et al.~\cite{Jones:2022io}, we do not employ the newly generated examples as synthetic positive instances (in our case: for the class $A$); otherwise, the classifier would learn to label fraudulent instances as $A$, which is not our \hl{goal}.

\begin{figure*}[h]
\centering
\includegraphics[width=.7\textwidth]{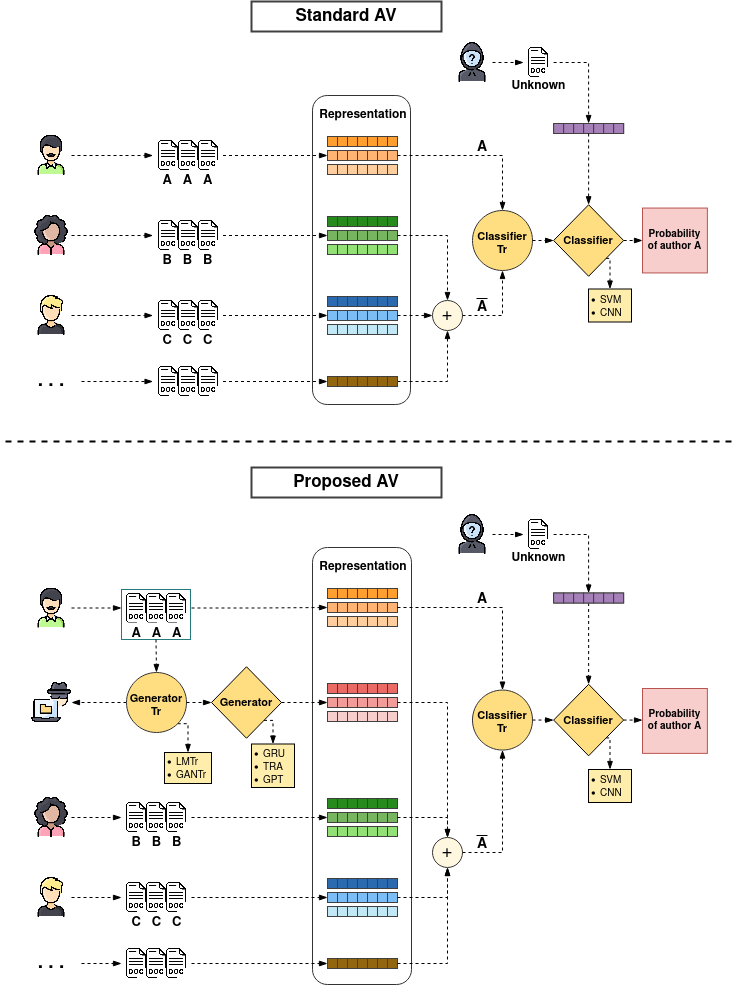}
\caption{Upper: Flowchart of a standard AV method. Bottom: Flowchart of our proposed AV method, where representative examples of forgery are added to $\overline{A}$.}
\label{fig:flowchart}
\end{figure*}

This process is sketched in Figure~\ref{fig:flowchart}.\footnote{Icons made by Vitaly Gorbachev on Flaticon: \url{https://www.flaticon.com/}}
As shown in the diagram, we explore three different generator architectures (\gru, \transformer, and \gpt) that we \hl{discuss} in Section~\ref{subsec:method_gen}, and that we train using two different learning algorithms (\LMtraining\ and \GANtraining) that we \hl{describe} in Section~\ref{subsec:method_training}. Concerning the underlying classifier of our AV system, we experiment with two learning algorithms (\svm\ and \nn) that we \hl{discuss} in Section~\ref{subsec:method_classifier}.

\subsection{Synthesizing forgery documents}
In this section, we describe the generation process by which we obtain new synthetic documents meant to imitate the production of $A$. In Section~\ref{subsec:method_gen} we describe the generator architectures that we explore in our experiments, while in Section~\ref{subsec:method_training} we describe the training strategies we employ.

\subsubsection{Generator architectures}
\label{subsec:method_gen}
\noindent In order to generate the adversarial examples, we experiment with three alternative generator architectures of increasing level of complexity:

\begin{itemize}
    \item \gru: Ezen~\cite{Ezen:2020no} \hl{shows} that simpler recurrent models (specifically: LSTM) tend to outperform more sophisticated models (specifically: BERT) when the training data is small, since simpler models have fewer parameters and are thus less prone to overfitting. Given that data scarcity is characteristic of many AV settings, we consider the Gated Recurrent Unit (GRU) \cite{Cho:2014ps} model, a simplified variant of LSTM. We set our model with 2 unidirectional GRU-layers of 512 hidden units each, followed by a linear layer with a ReLU activation function.\footnote{We use the PyTorch implementation of GRU: \url{https://pytorch.org/docs/stable/generated/torch.nn.GRU.html}}

    \item \transformer: We consider the original transformer introduced by Vaswani et al.~\cite{Vaswani:2017tt}, that replaced a recurrently-handled memory in favor of fully-attention layers.
    We set our model with 2 encoder layers of 512 hidden dimensions each and 4 attention heads, followed by a linear layer with a ReLU activation.\footnote{Our implementation relies on Pytorch's \texttt{TransformerEncoder}:  \url{https://pytorch.org/docs/stable/generated/torch.nn.TransformerEncoder.html}}
    
    \item \gpt: The forefront in current generation models is held by very large LMs, with Chat-GPT 4 occupying the top position. Unfortunately, these models are huge in terms of the number of parameters, and are typically not released to the scientific community, if not behind a pay-wall and an API. As a representative (though much smaller) model, we focus on GPT-2, a pre-trained unidirectional transformer \cite{Radford:2019uo} released by OpenAI that became increasingly popular for its outstanding generation performance and general-purpose nature. Uchendu et al.~\cite{Uchendu:2020pa} and Fagni et al.~\cite{Fagni:2021tf} assessed the quality of its generated texts, concluding \hl{that} these are often more human-like than those from other text generators. To limit the number of parameters and speed up the computation, while retaining the good quality of the original model, we employ DistilGPT2 \cite{Sanh:2019tb}, a smaller model (82M parameters versus the 124 M parameters of the standard version of GPT-2) that is trained via knowledge distillation on GPT-2. Note that this model outputs a multinomial distribution over the entire vocabulary at each generation step, and then selects the next token by using a sampling technique; we use the top-$k$ sampling, where the $k$ most-likely next words are filtered, and the probability mass is redistributed among those $k$ words (we set $k=50$). We enforce no repetitions within $5$-grams.\footnote{We \hl{rely} on the Huggingface's \texttt{transformers} implementation: \url{https://huggingface.co/distilgpt2}.} 
   
\end{itemize}

\noindent We explore different vector representations for the input and output of the generators: 

\begin{itemize}
    \item One-hot encoding (\oneh): a typical representation in which a vector of length $|V|$, where $V$ is the vocabulary, contains exactly one ``1'' whose index identifies one specific word in $V$ (all other values are set to ``0''). We use the DistilGPT2 word tokenizer in all our experiments, which \hl{results in} a vocabulary size of $|V|=$50,257 tokens. By \gruOnehot\ and \transformerOnehot\ we denote the variants equipped with an input linear (without bias) layer that projects the one-hot representation onto 128 dimensions, hence densifying the representation (GPT is a more complex model and we do not explore the \oneh\ variant). The output of these models is generated by a last linear layer of $|V|$ dimensions. During the \GANtraining, we compute the Gumbel-Softmax distribution over the output in order to discretize the output \hl{for} the next word in the sequence without interrupting the gradient.\footnote{We rely on the PyTorch implementation: \url{https://pytorch.org/docs/stable/generated/torch.nn.functional.gumbel_softmax.html}} 
    \item Dense encoding (\emb): we also experiment with a variant in which the generator produces embeddings of the same dimension of the ones used by the \nn\ method of Section~\ref{subsec:method_classifier}; we call the three models \gruEmbed, \transformerEmbed, and \gptEmbed, respectively. 
\end{itemize}

\subsubsection{Generator training}
\label{subsec:method_training}
\noindent We explore two different strategies for training the generator architectures described above: one based on standard Language-Model training (hereafter: \LMtraining), in which the generator is trained to replicate the style of the author of interest, and another based on GAN training (hereafter: \GANtraining), where the generator plays the role of \hl{a} forger that tries to fool the discriminator.

\textbf{Language Model Training (\LMtraining):}
\label{subsubsec:method_training_lm}
\noindent A standard way in which LMs are trained comes down to optimizing the model to predict the next word in a sequence, for a typically large number of sequences. More formally, given a sentence $w_1,\ldots,w_t$ of $t$ tokens, the model is trained to maximize the conditional probability $\Pr(w_{t}|w_{1}, w_{2} ... , w_{t-1};\Theta)$, where $\Theta$ are the model parameters. Such sequences are drawn from real textual examples, and could either come from a generic domain (thus optimizing a LM \hl{for} a language in general) or from a specific domain (thus optimizing the LM for a particular area of knowledge or task). 

We are interested in the latter case. Specifically, we draw sequences from texts that we know have been written by $A$, thus attaining a LM that tries to imitate the author (i.e., that tries to choose the next word as $A$ would have chosen).  
This idea has been shown to retain the stylistic patterns typical of the target author to a certain extent \cite{Jones:2022io}. Of course, the main limitation of this strategy is the limited amount of sequences we might expect to have access to, since these are bounded by the production of one single author; such sequences might be very few when compared with the typical amount of information used to train LMs. 

\textbf{Generative Adversarial Network Training (\GANtraining):}
\label{subsubsec:method_training_gan}
\noindent 
The Wasserstein GAN (WGAN) approach \cite{Arjovsky:2017ws} is based on a GAN architecture (see Section~\ref{sec:related_work}) that relies on the Earth-Mover (also called Wasserstein) distance as the loss function.\footnote{In the related literature, the discriminator underlying a WGAN is sometimes called ``the critic'' instead of ``the discriminator'', since it outputs a confidence score in place of a posterior probability. This distinction is rather unimportant for the scope of our paper, so we keep the term ``discriminator'' for the sake of simplicity.} This loss is continuous everywhere and produces smoother gradients, something that has been shown to prevent the gradient-vanishing problem and the mode-collapse problem that typically affect the early stages of the GAN training, when the generator $G$ still performs poorly. Gulrajani et al.~\cite{Gulrajani:2017di} later developed WGANGP, that improves the stability of the training by penalizing the gradient of the discriminator $D$ instead of clipping the weights (as proposed in the original formulation), which is the approach that we adopt in this paper.\footnote{We use the \texttt{PyTorch}-based implementation available at: \url{https://github.com/eriklindernoren/PyTorch-GAN/tree/master/implementations/wgan_gp}}

\hl{We leverage GAN training to generate samples that the classifiers find challenging to distinguish from the positive class (i.e., the author of interest). Incorporating these generated samples into the training set should enhance the classifier's robustness, improving its ability to differentiate the author's production from other instances (see Section\mbox{~\ref{sec:related_work}}).}

As the generator $G$, we explore the architectures (\gru, \transformer, \gpt) described in Section~\ref{subsec:method_gen}, while for the discriminator $D$ we employ a \nn-based classifier (the same classifier that we describe in detail in Section~\ref{subsubsec:cnn}).

\subsection{AV classifiers}
\label{subsec:method_classifier}
\noindent We experiment with two different classifiers for our AV system: one based on \svm\ (Section~\ref{subsubsec:svm}) and another based on \nn\ (Section~\ref{subsubsec:cnn}).
Before describing the learning algorithms we employ, we present in Section~\ref{subsubsec:basefeatures} \hl{the} set of features we extract that are commonly employed in the authorship analysis literature.

\subsubsection{Base Features}
\label{subsubsec:basefeatures}

\noindent The following set of features have proven useful in the related literature and \hl{are now considered} standard in many AId studies. We hereafter refer to this set as ``Base Features'' (BFs).

\begin{itemize}
    \item Function words: the normalized relative frequency of each function word. For a discussion about this type of features, see e.g., the analysis conducted by Kestemont~\cite{Kestemont:2014uc}. We use the list provided for the English stopwords by the NLTK library.\footnote{\url{https://www.nltk.org/}}
    \item Word lengths: the relative frequency of words up to a certain length, where a word length is the number of characters that forms a word. We set the range of word lengths to $[1,n]$, where $n$ is the longest word appearing at least 5 times in the training set. These are standard features employed in statistical authorship analysis since Mendenhall's ``characteristic curves of composition'' \cite{Mendenhall:1887cc}.
    \item POS-tags: the normalized relative frequency of each Part-Of-Speech (POS) tag. POS tags are an example of syntactic features, and are often employed in AId studies, also thanks to their topic-agnostic nature. We extract the POS-tags using the Spacy English tagger module.\footnote{ \url{https://spacy.io/usage/linguistic-features#pos-tagging}}
\end{itemize}

\subsubsection{SVM classifier}
\label{subsubsec:svm}

\noindent We consider a \svm-based classifier as our AV system, due to the good performance \svm s have demonstrated in text classification tasks in general over the years \cite{Joachims98}, and in authorship analysis\hl{-}related tasks in particular \cite{Kestemont:2019wo}.

For \svm, we employ the implementation from the \texttt{scikit-learn} package.\footnote{\url{https://scikit-learn.org/stable/modules/generated/sklearn.svm.SVC.html}} 
We optimize the hyper-parameters of the classifier via grid-search. In particular, we explore the parameter $C$ in the range $[0.001, 0.01, 0.1, 1, 10, 100, 1000]$, the parameter kernel in the range $\{$\textit{linear}, \textit{poly}, \textit{rbf}, \textit{sigmoid}$\}$, and we explore whether to rebalance the class weights or not. After model selection, the \svm\ is then re-trained on the union of training and validation sets with the optimized hyper-parameters, and then evaluated on the test set. 

\hl{We use the BFs set as features.}

\subsubsection{CNN classifier}
\label{subsubsec:cnn}

\begin{figure*}[ht]
\centering
\includegraphics[width=.95\textwidth]{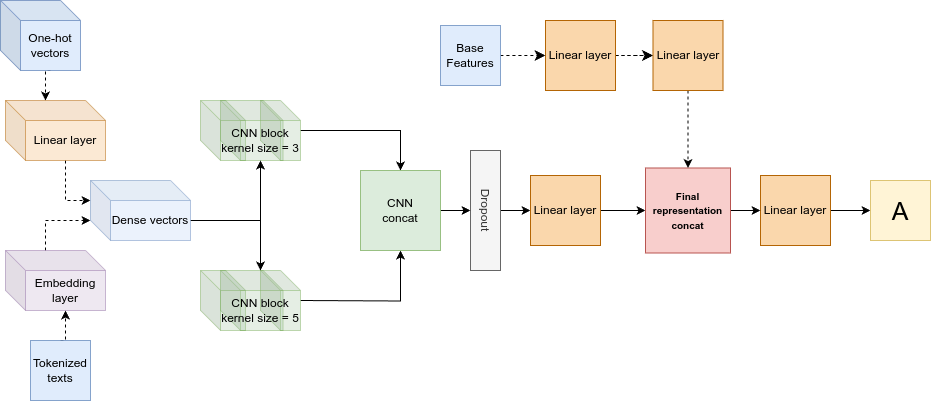}
\caption{The \nn\ classifier architecture. The dotted lines represent alternative branches.
}
\label{fig:NN_classifier}
\end{figure*}

\noindent %The classifier we employ for our discriminator $D$ in the \GANtraining, that we here reuse as the underlying classifier of our AV system, is based on a \nn\ architecture.
We also experiment with an AV classifier based on the following \nn\ architecture. The input layer of the \nn\ has variable dimensions depending on the vector modality: the dense representations (\emb) depend on the generator architecture of choice (\gpt\ produces 768-dimensional vectors, while we set the output dimensions of \gru\ and \transformer\ to 128); while the one-hot representations (\oneh) consist of $|V|=$50,257 dimensions that are projected into a 128-dimensional space by means of a linear layer (without bias). This is followed by two parallel convolutional blocks with kernel sizes of 3 and 5, respectively. Each convolutional block consists of two layers of 512 and 256 dimensions and ReLU activations. We then apply max-pooling to the resulting tensors and concatenate the outputs of both blocks. We also apply dropout with 0.3 probability, and the resulting tensor is then processed by a linear transformation of 64 dimensions with ReLU activation. For generator architectures for which we can derive ``plain texts'' (i.e., for all the \oneh-variants plus \gpt), we add a parallel branch that receives their BFs as inputs: these features are processed by two linear layers of 128 and 64 dimensions and ReLU activations. The resulting tensor is then concatenated with the output of the other branch, and the result is passed through a final linear layer with ReLU activation which produces a single value representing the confidence score of the classifier. The complete \nn\ model is depicted in Figure~\ref{fig:NN_classifier}.
 
We train the network using the AdamW optimizer \cite{Loshchilov:2018dd} with a learning rate of $0.001$, a batch size of 32, and binary cross entropy as the loss function. In order to counter the effect of class imbalance (there are many more negatives than positives), we set the class weights to the ratio between negative and positive examples.

We train the model for a minimum of 50 epochs and a maximum of 500 epochs, and apply early stopping when the performance on the validation set (as measured in terms of $F_1$) does not improve for 25 consecutive epochs. Finally, we re-train the model on the combination of training and validation sets for 5 epochs before the model evaluation.

% -----------------------------------------------------

\section{Experimental setting}
\label{sec:exp}
\noindent In this section, we present the experiments we have carried out. In Section~\ref{subsec:exp_dataset} we describe the datasets we employ, while in Section~\ref{subsec:exp_prot} we describe the experimental protocol we follow. Finally, Section~\ref{subsec:exp_results} discusses the results we have obtained.

All the models and experiments are developed in Python, employing the \texttt{scikit-learn} library \cite{Pedregosa:2011nn} and the \texttt{PyTorch} library \cite{Paszke:2019th}. The code to reproduce all our experiments is available on GitHub.\footnote{\url{https://github.com/silvia-cor/Authorship_DataAugmentation}}

\subsection{Datasets}
\label{subsec:exp_dataset}
\noindent We experiment with five publicly available datasets, some of which are examples of a close-set setting (where the authors comprising the test set are also present in the training set), while others are instead open-set (where the authors comprising the test set are not necessarily present in the training set).

\begin{itemize}

  \item \texttt{\textbf{TweepFake}}. This dataset was created and made publicly available by Fagni et al.~\cite{Fagni:2021tf};\footnote{A limited version is available on Kaggle at: \url{https://www.kaggle.com/datasets/mtesconi/twitter-deep-fake-text}}\hl{ it contains} tweets from 17 human accounts and 23 bots, each one imitating one of the human accounts. The dataset is balanced (the tweets are half human- and half bot- generated) and already partitioned into a training, a validation, and a test set. Since it would realistically be rather problematic to obtain a corpus containing human forgeries, we use this dataset as a reasonable proxy, where the forgery is made by a machine trying to emulate a human writer. In order to reproduce a more realistic setting, we only consider the documents produced by human users for our training and validation sets, but we keep all the documents in the test set \hl{in order} to emulate an open-set problem.
  
  \item \texttt{\textbf{EBG}}. The Extended Brennan-Greenstadt Corpus was created by Brennan et al.~\cite{Brennan:2012la}; we use the Obfuscation setting\footnote{ 
  Available on the Reproducible Authorship Attribution Benchmark Tasks (RAABT) on Zenodo: \url{https://zenodo.org/record/5213898##.YuuaNdJBzys}\label{EBG_corpus}} consisting of writings from 45 individuals contacted through the Amazon Mechanical Turk platform. Participants were asked to: i) upload examples of their own writing (of ``scholarly'' nature), and ii) to write a short essay regarding the description of their neighborhood while obscuring their writing style, without any specific instructions on how to do so. We randomly select 10 authors and use all their documents in i) to compose the train-and-validation set; we split it with a $90/10$ ratio into training and validation sets, in a stratified fashion. All the documents in ii) form the test set, making it an open-set scenario. We use these documents in order to check the ability of the model to recognize the author of interest even in cases where they actively try to mask their writing.
  
  \item \texttt{\textbf{RJ}}. The Riddell-Juola Corpus was created by Riddell et al.~\cite{Riddell:2021ll};\footnote{Available on the same link of the EBG corpus at Footnote~\ref{EBG_corpus}.} we use the Obfuscation setting. The documents were collected with the same policy as in the EBG corpus, and we use them for analogous reasons. In this corpus the participants were randomly assigned to receive the instruction to obfuscate their writing style. We randomly select 10 authors and split the train-and-validation set as in the EBG corpus, while keeping the whole test set, thus making it an example of an open-set scenario.
  
  \item \texttt{\textbf{PAN11}}. This dataset is based on the Enron email corpus\footnote{Available at: \url{http://www.cs.cmu.edu/~enron/}} and \hl{was} developed for the PAN2011 authorship competition \cite{Argamon:2011ew}.\footnote{Available at: \url{https://pan.webis.de/clef11/pan11-web/authorship-attribution.html}} It contains three distinct problem settings (each coming with its own training, validation, and test set) for AV, where each training set contains emails by a single author, and each validation and test set contains a \hl{mix} of documents written by \hl{both} the training author and others (not all from the Enron corpus). Personal names and email addresses in the corpus have been redacted; moreover, in order to reflect a \hl{real-world} task environment, some texts are not in English, or are automatically generated. We merge the three training sets in a single one (resulting in a training set with three authors), and we do the same for the validation and test sets. We use this dataset as an example of contemporary production ``in the wild'', where the authors are realistically not trying to conceal their style, and are prone to all the mistakes and noise of digital communication. Moreover, the training set is rather limited (there are only two authors representing the $\overline{A}$ class), and the resulting AV problems are open-set.
  
    \item \texttt{\textbf{Victoria}}. This dataset was created and made publicly available by Gungor~\cite{Gungor:2018hv}.\footnote{Available at: \url{https://archive.ics.uci.edu/ml/datasets/Victorian+Era+Authorship+Attribution}. We only employ the `train' dataset, since the `test' dataset does not contain the authors' labels.} It consists of books by American or English 18th-19th century novelists, divided into segments of 1,000 words each. The first and last 500 words of each book have not been included, and only the 10,000 most frequent words have been retained, while the rest have been deleted. The result is a corpus of more than 50,000 documents by 50 different authors. We use these documents as examples of literary production, where no author is presumably trying to imitate someone else's style, nor conceal their own. We limit the dataset to 5 authors selected randomly with 1,000 chunks each (see below), in order to \hl{address} the experimental setting where, unlike in the other datasets, there are few authors but a \hl{substantial amount} of data. We divide the data into a training-and-validation set and a test set with a $90/10$ ratio, and further divide the former into a training and a validation set with another $90/10$ ratio, in a stratified fashion. This dataset is representative of a closed-set AV problem.
\end{itemize}

\noindent In each dataset, we split each document into non-overlapping chunks of 100 tokens (words and punctuation), and we discard \hl{the} chunks of less than 25 words (not considering punctuation); we carry out this segmentation before partitioning the dataset into a training, validation, and test set. We also exclude authors with less than 10 chunks in the training set. Table~\ref{tab:dataset} shows the final number of training, validation, and test examples for each dataset.

\begin{table}[ht]
\centering
\caption{Number of authors in training, and number of training, validation, and test examples in each dataset.}
\label{tab:dataset}
\begin{tabular}{l|crrr|}
\cline{2-5}
\multicolumn{1}{r|}{} & \multicolumn{1}{c}{\textbf{\#authors}} & \multicolumn{1}{c}{\textbf{\#training}} & \multicolumn{1}{c}{\textbf{\#validation}} & \multicolumn{1}{c|}{\textbf{\#test}} \\ \hline
\multicolumn{1}{|l|}{\texttt{\textbf{TweepFake}}} & 15 & 3,099 & 331 & 761  \\ \hline
\multicolumn{1}{|l|}{\texttt{\textbf{EBG}}} & 10 & 800 & 89 & 270 \\ \hline
\multicolumn{1}{|l|}{\texttt{\textbf{RJ}}} & 10 & 598 & 67 & 161 \\ \hline
\multicolumn{1}{|l|}{\texttt{\textbf{PAN11}}} & 3 & 90 & 47 & 348 \\ \hline
\multicolumn{1}{|l|}{\texttt{\textbf{Victoria}}} & 5 & 4,050 & 450 & 500 \\ \hline
\end{tabular}%
\end{table}

\subsection{Experimental protocol}
\label{subsec:exp_prot}

\noindent For each dataset, \hl{we conduct experiments in rounds, treating} each author as the positive class $A$, while the remaining authors \hl{serve} as the negative class $\overline{A}$.

\hl{At} each generation step, we generate $n$ new examples, where $n$ is set to 10 times the number of training chunks for $A$, up to a maximum of 1,000 new generated examples. As prompt for each new generation, we use the first 5 tokens from a randomly selected training chunk by $A$; the generated text has the same length as the original chunk by $A$. Once the $n$ examples have been generated, they are labelled as $\overline{A}$ and added to the training set, which is used to train the classifiers discussed in Section~\ref{subsec:method_classifier}.

We denote the classifiers trained with data augmentation with the following nomenclature: $C+G^{T}_{E}$, where $C$ is a classifier from Section~\ref{subsec:method_classifier}, $G$ is a generator architecture from Section~\ref{subsec:method_gen}, $T$ is a generator training strategy from Section~\ref{subsec:method_training}, and $E$ is an encoding type (\oneh\ or \emb\ --- see Section~\ref{subsec:method_gen}). Additional details specific to each combination are reported in the Appendix. Note that the natural baseline for any augmentation setup $C+G^{T}_{E}$ is $C$, i.e., the same classifier trained without the newly generated examples.

We measure the performance of the classifier in terms of: i) the well-known $F_{1}$ metric, given by:
\begin{equation*}
    F_1=\left\{\begin{matrix}
\frac{2\mathrm{TP}}{2\mathrm{TP}+\mathrm{FP}+\mathrm{FN}} & \mathrm{if}\; (\mathrm{TP}+\mathrm{FP}+\mathrm{FN})>0\\ 
1 & \mathrm{otherwise}
\end{matrix}\right.
\end{equation*}

\noindent where $\mathrm{TP}, \mathrm{FP}$, and $\mathrm{FN}$ stand for the number of true positive, false positives, and false negatives, respectively,\footnote{\label{foot:f1}Note that the $F_{1}$ metric does not take into account the true negatives. We set $F_{1}=1$ if there are no true positives and the classifier guesses all the true negatives correctly.}, and ii) in terms of the $K$ metric \cite{Sebastiani:2015xl} given by:
\begin{equation*}
\footnotesize
    K=\left\{\begin{matrix}
\frac{TP}{TP+FN}+\frac{TN}{TN+FP}-1 & \mathrm{if} & (TP+FN>0)\land (TN+FP>0) \\ 
2\cdot\frac{TN}{TN+FP}-1 & \mathrm{if} & (TP+FN=0)\\ 
2\cdot\frac{TP}{TP+FN}-1 & \mathrm{if} & (TN+FP=0)
\end{matrix}\right.
\end{equation*}
\normalsize

\noindent where $\mathrm{TN}$ stand for the number of true negatives. Note that $F_1$ ranges from 0 (worst) to 1 (best), while $K$ ranges from -1 (worst) to 1 (best), with 0 corresponding to the accuracy of the random classifier. We report the average in performance, both in terms of $F_1$ and in terms of $K$, across all experiments per dataset. 

Since our goal is to improve the classifier performance \hl{ by adding synthetically generated examples}, we compute the relative improvement of the classifier trained on the augmented dataset compared to the same classifier trained \hl{exclusively} on the original training set (excluding adversarial examples).
We also compute the statistical significance of the differences in performance via the McNemar's paired non-parametric statistical hypothesis test \cite{McNemar:1947jh}. To this aim, we convert the outputs of the two methods (with and without augmentation) into values $1$ (correct prediction) and $0$ (wrong prediction). We take $0.05$ as the confidence value for statistical significance.

\subsection{Results}
\label{subsec:exp_results}

Table~\ref{tab:results4} reports the results for the different combinations of classifiers and generation procedures. Note that \svm\ \hl{is} not combined with the \emb-variants of \gru\ or \transformer\ since, in those cases, and in contrast to \gpt, we are not able to generate plain texts from which BFs can be extracted. 

\renewcommand{\arraystretch}{2}
\begin{table*}
\centering
\Large
\caption{Results of our experiments with the \svm\ and \nn\ learning algorithms. Groups of experiments sharing the same learning algorithm, with and without data augmentation from the various generators, are reported on two consecutive rows. For each experiment we report: the values of $F_{1}$ and $K$, the percentage of improvement (\textbf{$\Delta\%$}) resulting from the addition of generated data, and the results of the McNemar statistical significance test (\textbf{M}) against the baseline (\cmark : statistical significance confirmed; \xmark : statistical significance rejected). The best result obtained for the given dataset and evaluation measure for each experiment \hl{group} is in \textbf{bold}, while the worst is in \textit{italic}; the best result obtained for the given dataset and evaluation measure overall is in \underline{\textbf{underlined bold}}. Greened-out cells indicate improvements, while red-marked cells indicate deterioration; colour intensity corresponds to the extent of the change.
}
\label{tab:results4}
\resizebox{\textwidth}{!}{%
\setlength{\tabcolsep}{3pt}
\begin{tabular}{lrrrrc|rrrrc|rrrrc|rrrrc|rrrrc}
\toprule

 & \multicolumn{5}{c|}{\textbf{TweepFake}} & \multicolumn{5}{c|}{\textbf{EBG}} & \multicolumn{5}{c|}{\textbf{RJ}} & \multicolumn{5}{c|}{\textbf{PAN11}} & \multicolumn{5}{c}{\textbf{Victoria}} \\ \cmidrule(rl){2-6} \cmidrule(rl){7-11} \cmidrule(rl){12-16} \cmidrule(rl){17-21} \cmidrule(rl){22-26}

 & \multicolumn{1}{c}{\textbf{$F_{1}$}} & \multicolumn{1}{c}{\textbf{$\Delta\%$}} & \multicolumn{1}{c}{\textbf{$K$}} & \multicolumn{1}{c}{\textbf{$\Delta\%$}} & \textbf{M} & \multicolumn{1}{c}{\textbf{$F_{1}$}} & \multicolumn{1}{c}{\textbf{$\Delta\%$}} & \multicolumn{1}{c}{\textbf{$K$}} & \multicolumn{1}{c}{\textbf{$\Delta\%$}} & \textbf{M} & \multicolumn{1}{c}{\textbf{$F_{1}$}} & \multicolumn{1}{c}{\textbf{$\Delta\%$}} & \multicolumn{1}{c}{\textbf{$K$}} & \multicolumn{1}{c}{\textbf{$\Delta\%$}} & \textbf{M} & \multicolumn{1}{c}{\textbf{$F_{1}$}} & \multicolumn{1}{c}{\textbf{$\Delta\%$}} & \multicolumn{1}{c}{\textbf{$K$}} & \multicolumn{1}{c}{\textbf{$\Delta\%$}} & \textbf{M} & \multicolumn{1}{c}{\textbf{$F_{1}$}} & \multicolumn{1}{c}{\textbf{$\Delta\%$}} & \multicolumn{1}{c}{\textbf{$K$}} & \multicolumn{1}{c}{\textbf{$\Delta\%$}} & \textbf{M} \\ \midrule

\textbf{\svm} & .366 & & .375 & & & \textbf{.455} & & .038 &  & & \underline{\textbf{.621}} & & .296 &  & & .280 & & .242 &  & & .673 & & .606 & & \\

\textbf{\svmLMonehotGru} & \cellcolor{red!42}.335 & \cellcolor{red!42}$-8.48$ & \cellcolor{green!4}.378 & \cellcolor{green!4}$+0.83$ & \cmark & \cellcolor{red!30}.427 & \cellcolor{red!30}$-6.07$ & \cellcolor{green!50}\textbf{.087} & \cellcolor{green!50}$+129.06$ & \cmark & \cellcolor{red!50}.524 & \cellcolor{red!50}$-15.55$ & \cellcolor{green!1}.296 & \cellcolor{green!1}$+0.24$ & \xmark & \cellcolor{red!50}.252 & \cellcolor{red!50}$-10.00$ & \cellcolor{red!50}\textit{.088} & \cellcolor{red!50}$-63.55$ & \cmark & \cellcolor{green!0}.673 & \cellcolor{green!0}$+0.06$ & .606 & 0.00 & \xmark \\

\textbf{\svmGANonehotGru} & \cellcolor{green!26}\textbf{.385} & \cellcolor{green!26}$+5.34$ & \cellcolor{red!41}\textit{.344} & \cellcolor{red!41}$-8.21$ & \cmark & \cellcolor{red!29}.428 & \cellcolor{red!29}$-5.87$ & \cellcolor{red!24}.036 & \cellcolor{red!24}$-4.97$ & \cmark & \cellcolor{red!50}.530 & \cellcolor{red!50}$-14.75$ & \cellcolor{green!41}\underline{\textbf{.320}} & \cellcolor{green!41}$+8.22$ & \xmark & \cellcolor{red!50}.185 & \cellcolor{red!50}$-34.05$ & \cellcolor{green!26}\textbf{.255} & \cellcolor{green!26}$+5.36$ & \cmark & \cellcolor{red!3}.668 & \cellcolor{red!3}$-0.62$ & \cellcolor{green!2}.609 & \cellcolor{green!2}$+0.40$ & \xmark \\

\textbf{\svmLMonehotTrans} & \cellcolor{red!50}\textit{.324} & \cellcolor{red!50}$-11.39$ & \cellcolor{green!18}.389 & \cellcolor{green!18}$+3.77$ & \cmark & \cellcolor{red!26}.430 & \cellcolor{red!26}$-5.36$ & \cellcolor{green!50}.074 & \cellcolor{green!50}$+93.19$ & \cmark & \cellcolor{red!50}.430 & \cellcolor{red!50}$-30.77$ & \cellcolor{red!5}.292 & \cellcolor{red!5}$-1.08$ & \xmark & \cellcolor{red!17}.270 & \cellcolor{red!17}$-3.45$ & \cellcolor{red!50}.160 & \cellcolor{red!50}$-34.11$ & \cmark & \cellcolor{red!5}.665 & \cellcolor{red!5}$-1.07$ & \cellcolor{red!2}.604 & \cellcolor{red!2}$-0.40$ & \cmark \\

\textbf{\svmGANonehotTrans} & \cellcolor{red!6}.361 & \cellcolor{red!6}$-1.20$ & \cellcolor{red!18}.361 & \cellcolor{red!18}$-3.77$ & \cmark & \cellcolor{red!37}\textit{.421} & \cellcolor{red!37}$-7.45$ & \cellcolor{red!50}.020 & \cellcolor{red!50}$-47.64$ & \cmark & \cellcolor{red!50}.526 & \cellcolor{red!50}$-15.34$ & \cellcolor{green!21}.308 & \cellcolor{green!21}$+4.30$ & \cmark & \cellcolor{red!50}\textit{.182} & \cellcolor{red!50}$-35.12$ & \cellcolor{red!50}.197 & \cellcolor{red!50}$-18.71$ & \cmark & \cellcolor{red!5}.666 & \cellcolor{red!5}$-1.04$ & \cellcolor{red!9}\textit{.595} & \cellcolor{red!9}$-1.91$ & \xmark \\

\textbf{\svmLMgpt} & \cellcolor{green!5}.369 & \cellcolor{green!5}$+1.02$ & \cellcolor{green!47}\underline{\textbf{.411}} & \cellcolor{green!47}$+9.49$ & \cmark & \cellcolor{red!31}.426 & \cellcolor{red!31}$-6.24$ & \cellcolor{red!50}\textit{.003} & \cellcolor{red!50}$-93.19$ & \cmark & \cellcolor{red!50}\textit{.311} & \cellcolor{red!50}$-49.98$ & \cellcolor{red!50}\textit{.246} & \cellcolor{red!50}$-16.88$ & \xmark & \cellcolor{green!50}\textbf{.399} & \cellcolor{green!50}$+42.38$ & \cellcolor{red!50}.159 & \cellcolor{red!50}$-34.39$ & \cmark & \cellcolor{green!2}\textbf{.676} & \cellcolor{green!2}$+0.48$ & \cellcolor{green!21}\textbf{.632} & \cellcolor{green!21}$+4.22$ & \xmark \\

\textbf{\svmGANgpt} & \cellcolor{red!49}.330 & \cellcolor{red!49}$-9.83$ & \cellcolor{red!1}.374 & \cellcolor{red!1}$-0.32$ & \cmark & \cellcolor{red!15}.441 & \cellcolor{red!15}$-3.17$ & \cellcolor{red!50}.028 & \cellcolor{red!50}$-26.96$ & \cmark & \cellcolor{red!50}.517 & \cellcolor{red!50}$-16.73$ & \cellcolor{red!50}.264 & \cellcolor{red!50}$-10.52$ & \cmark & \cellcolor{green!50}.393 & \cellcolor{green!50}$+40.36$ &  \cellcolor{red!50}.136 & \cellcolor{red!50}$-43.74$ & \cmark & \cellcolor{red!6}\textit{.664} & \cellcolor{red!6}$-1.28$ & \cellcolor{red!7}.598 & \cellcolor{red!7}$-1.45$ & \xmark \\

 \hline

\textbf{\nnonehot$_{(\oneh)}$} & .617 &  & .376 &  & & .622 &  & \textit{.022} &  & & .326 &  & \textit{.230} &  & & \textbf{.331} &  & \textbf{.407} &  & & .756 &  & .655 &  & \\

\textbf{\nnLMonehotGru} & \cellcolor{red!27}.583 & \cellcolor{red!27}$-5.47$ & \cellcolor{red!28}\textit{.354} & \cellcolor{red!28}$-5.68$ & \cmark & \cellcolor{red!44}.566 & \cellcolor{red!44}$-8.90$ & \cellcolor{green!50}.072 & \cellcolor{green!50}$+227.27$ & \xmark & \cellcolor{red!50}\textit{.238} & \cellcolor{red!50}$-26.98$ & \cellcolor{green!50}\textbf{.313} & \cellcolor{green!50}$+35.94$ & \cmark & \cellcolor{red!50}.263 & \cellcolor{red!50}$-20.72$ & \cellcolor{red!4}.403 & \cellcolor{red!4}$-0.90$ & \cmark & \cellcolor{green!7}\underline{\textbf{.767}} & \cellcolor{green!7}$+1.51$ & \cellcolor{green!19}\underline{\textbf{.681}} & \cellcolor{green!19}$+3.88$ & \xmark \\

\textbf{\nnGANonehotGru} & \cellcolor{red!50}.526 & \cellcolor{red!50}$-14.68$ & \cellcolor{red!26}.356 & \cellcolor{red!26}$-5.20$ & \xmark & \cellcolor{red!50}\textit{.444} & \cellcolor{red!50}$-28.59$ & \cellcolor{green!50}.117 & \cellcolor{green!50}$+430.00$ & \xmark & \cellcolor{green!50}\textbf{.521} & \cellcolor{green!50}$+59.63$ & \cellcolor{green!50}.285 & \cellcolor{green!50}$+24.03$ & \cmark & \cellcolor{red!50}\textit{.175} & \cellcolor{red!50}$-47.08$ & \cellcolor{red!50}\textit{.349} & \cellcolor{red!50}$-14.25$ & \cmark & \cellcolor{red!5}.747 & \cellcolor{red!5}$-1.09$ & \cellcolor{red!5}.648 & \cellcolor{red!5}$-1.07$ & \xmark \\

\textbf{\nnLMonehotTrans} & \cellcolor{red!50}\textit{.363} & \cellcolor{red!50}$-41.05$ & \cellcolor{green!15}\textbf{.387} & \cellcolor{green!15}$+3.03$ & \cmark & \cellcolor{red!50}.532 & \cellcolor{red!50}$-14.37$ & \cellcolor{green!50}.108 & \cellcolor{green!50}$+390.00$ & \cmark & \cellcolor{green!12}.334 & \cellcolor{green!12}$+2.51$ & \cellcolor{green!50}.304 & \cellcolor{green!50}$+32.16$ & \cmark & \cellcolor{red!50}.256 & \cellcolor{red!50}$-22.74$ & \cellcolor{red!50}.350 & \cellcolor{red!50}$-14.09$ & \cmark & \cellcolor{red!9}.742 & \cellcolor{red!9}$-1.83$ & \cellcolor{red!19}.629 & \cellcolor{red!19}$-3.97$ & \xmark \\

\textbf{\nnGANonehotTrans} & \cellcolor{green!7}\underline{\textbf{.626}} & \cellcolor{green!7}$+1.51$ & \cellcolor{green!7}.381 & \cellcolor{green!7}$+1.47$ & \xmark & \cellcolor{green!50}\textbf{.686} & \cellcolor{green!50}$+10.39$ & \cellcolor{green!50}\underline{\textbf{.163}} & \cellcolor{green!50}$+640.00$ & \cmark & \cellcolor{red!15}.316 & \cellcolor{red!15}$-3.13$ & \cellcolor{green!50}.291 & \cellcolor{green!50}$+26.64$ & \cmark & \cellcolor{red!50}.197 & \cellcolor{red!50}$-40.64$ & \cellcolor{red!0}.406 & \cellcolor{red!0}$-0.16$ & \cmark & \cellcolor{red!16}\textit{.731} & \cellcolor{red!16}$-3.28$ & \cellcolor{red!28}\textit{.618} & \cellcolor{red!28}$-5.71$ & \xmark \\

\hline

\textbf{\nnembed$_{(\#\emb=128)}$} & \textbf{.622} &  & \textit{.374} &  & & \textit{.427} &  & \textit{.022} &  & & .406 &  & .287 &  & & \textbf{.205} &  & \textbf{.371} &  & & .733 &  & .632 &  & \\

\textbf{\nnLMembedGru} & \cellcolor{red!50}.530 & \cellcolor{red!50}$-14.72$ & \cellcolor{green!27}.394 & \cellcolor{green!27}$+5.51$ & \cmark & \cellcolor{green!50}.628 & \cellcolor{green!50}$+47.03$ & \cellcolor{green!50}.065 & \cellcolor{green!50}$+188.84$ & \cmark & \cellcolor{red!50}\textit{.245} & \cellcolor{red!50}$-39.72$ & \cellcolor{red!12}\textit{.279} & \cellcolor{red!12}$-2.44$ & \cmark & \cellcolor{red!50}.177 & \cellcolor{red!50}$-13.68$ & \cellcolor{red!50}.319 & \cellcolor{red!50}$-13.94$ & \xmark & \cellcolor{red!2}.730 & \cellcolor{red!2}$-0.44$ & \cellcolor{green!19}\textbf{.657} & \cellcolor{green!19}$+3.89$ & \xmark \\

\textbf{\nnGANembedGru} & \cellcolor{red!50}.470 & \cellcolor{red!50}$-24.35$ & \cellcolor{green!42}\textbf{.406} & \cellcolor{green!42}$+8.53$ & \cmark & \cellcolor{green!50}.564 & \cellcolor{green!50}$+31.94$ & \cellcolor{green!50}\textbf{.101} & \cellcolor{green!50}$+352.23$ & \cmark & \cellcolor{green!29}.430 & \cellcolor{green!29}$+5.84$ & \cellcolor{red!3}.284 & \cellcolor{red!3}$-0.70$ & \cmark & \cellcolor{red!50}\textit{.174} & \cellcolor{red!50}$-14.82$ & \cellcolor{red!50}.315 & \cellcolor{red!50}$-14.93$ & \cmark & \cellcolor{red!27}\textit{.693} & \cellcolor{red!27}$-5.56$ & \cellcolor{red!22}\textit{.604} & \cellcolor{red!22}$-4.49$ & \cmark \\ 

\textbf{\nnLMembedTrans} & \cellcolor{red!50}\textit{.318} & \cellcolor{red!50}$-48.78$ & \cellcolor{green!40}.404 & \cellcolor{green!40}$+8.11$ & \cmark & \cellcolor{green!50}\textbf{.632} & \cellcolor{green!50}$+47.92$ & \cellcolor{green!50}.073 & \cellcolor{green!50}$+228.12$ & \cmark & \cellcolor{red!50}.249 & \cellcolor{red!50}$-38.56$ & \cellcolor{green!44}\textbf{.312} & \cellcolor{green!44}$+8.83$ & \cmark & \cellcolor{red!50}\textit{.174} & \cellcolor{red!50}$-14.98$ & \cellcolor{red!50}\textit{.280} & \cellcolor{red!50}$-24.46$ & \cmark & \cellcolor{red!15}.711 & \cellcolor{red!15}$-3.05$ & \cellcolor{red!16}.612 & \cellcolor{red!16}$-3.29$ & \cmark \\

\textbf{\nnGANembedTrans} & \cellcolor{red!50}.391 & \cellcolor{red!50}$-37.04$ & \cellcolor{green!26}.394 & \cellcolor{green!26}$+5.37$ & \cmark & \cellcolor{green!50}.629 & \cellcolor{green!50}$+47.29$ & \cellcolor{green!50}.065 & \cellcolor{green!50}$+189.29$ & \cmark & \cellcolor{green!50}\textbf{.539} & \cellcolor{green!50}$+32.68$ & .287 & 0.00 & \cmark & \cellcolor{red!50}.182 & \cellcolor{red!50}$-10.91$ & \cellcolor{red!47}.335 & \cellcolor{red!47}$-9.53$ & \xmark & \cellcolor{green!2}\textbf{.737} & \cellcolor{green!2}$+0.44$ & \cellcolor{green!14}.651 & \cellcolor{green!14}$+2.88$ & \xmark \\

\hline

\textbf{\nngpt$_{(\#\emb=768)}$} & \textbf{.482} &  & \textbf{.392} &  & & \textit{.451} &  & \textbf{.085} &  & & .513 &  & .301 &  & & \textit{.205} &  & .330 &  & & \textbf{.750} &  & .661 &  & \\

\textbf{\nnLMgpt} & \cellcolor{red!50}.418 & \cellcolor{red!50}$-13.26$ & \cellcolor{red!44}.356 & \cellcolor{red!44}$-8.96$ & \cmark & \cellcolor{green!50}\underline{\textbf{.810}} & \cellcolor{green!50}$+79.63$ & \cellcolor{red!50}\textit{-.005} & \cellcolor{red!50}$-105.55$ & \cmark & \cellcolor{red!0}\textit{.512} & \cellcolor{red!0}$-0.06$ & \cellcolor{red!47}\textit{.273} & \cellcolor{red!47}$-9.52$ & \xmark & \cellcolor{green!50}.362 & \cellcolor{green!50}$+76.14$ & \cellcolor{green!50}\underline{\textbf{.433}} & \cellcolor{green!50}$+31.08$ & \cmark & \cellcolor{red!25}\textit{.711} & \cellcolor{red!25}$-5.17$ & \cellcolor{red!50}\textit{.590} & \cellcolor{red!50}$-10.80$ & \xmark \\

\textbf{\nnGANgpt} & \cellcolor{red!50}\textit{.400} & \cellcolor{red!50}$-17.13$ & \cellcolor{red!50}\textit{.346} & \cellcolor{red!50}$-11.51$ & \xmark & \cellcolor{green!50}.619 & \cellcolor{green!50}$+37.26$ & \cellcolor{red!50}.006 & \cellcolor{red!50}$-92.44$ & \cmark & \cellcolor{green!50}\textbf{.619} & \cellcolor{green!50}$+20.71$ & \cellcolor{green!30}\underline{\textbf{.320}} & \cellcolor{green!30}$+6.10$ & \cmark & \cellcolor{green!50}\underline{\textbf{.564}} & \cellcolor{green!50}$+174.84$ & \cellcolor{red!50}\textit{.232} & \cellcolor{red!50}$-29.87$ & \cmark & \cellcolor{green!0}\textbf{.750} & \cellcolor{green!0}$+0.05$ & \cellcolor{green!4}\textbf{.667} & \cellcolor{green!4}$+0.91$ & \xmark \\ 

\bottomrule 

\end{tabular}%
}
\end{table*}

Statistical tests of significance reveal that data augmentation yields significant improvements in performance (for both metrics) in 11 out of 80 $\langle$method, dataset$\rangle$ combinations,  while it results in a deterioration in performance in 22 out of 80 combinations. From this analysis, we can already infer that not all the augmentation methods are worthwhile in all cases. 

A closer look reveals that augmentation has little impact, if at all, in the \texttt{Victoria} dataset; \hl{this was expected, given that this dataset has a large amount of original texts available, making synthetic augmentation likely to add only unnecessary second-hand information.} %\footnote{Given this, we exclude the \texttt{Victoria} dataset from the following analysis regarding the performance of the various methods with or without data augmentation.}. 
%However, beyond this, discriminating among those augmentation methods that positively pay off and those that do not, or even deducing a general ``rule-of-thumb'' by which to select the ``right'' method for the given dataset, appears tricky at best. 
Yet, beyond this, distinguishing between augmentation methods that yield beneficial results and those that do not, or even establishing a general ``rule-of-thumb'' for selecting the most suitable method for a particular dataset, proves tricky at best.

Indeed, it is rather difficult to pinpoint a single overall winner \hl{among} the various methods, with or without augmentation: out of 8 $\langle$evaluation measure, dataset$\rangle$ combinations, the methods equipping a \GANtraining\ augmentation \hl{result} in the best overall method 4 times, the methods equipping a \LMtraining\ augmentation 4 times, and the classifiers without augmentation only one time. In a more fine-grained view, out of 32 $\langle$method group, evaluation metric, dataset$\rangle$ combinations, the \GANtraining\ augmentation \hl{results} in the best method 13 times (11 of which are statistically significant), the \LMtraining\ augmentation 9 times (all statistically significant), and the classifiers without augmentation 10 times. Hence, while it might appear that the \GANtraining\ augmentation has a beneficial impact on \hl{the} classifier training in numerous cases, the positive effect is neither frequent nor consistent enough to yield definitive conclusions.

Therefore, no single \hl{method} appears to clearly outperform the others. It is worth noting, \hl{though}, that the \nn\ classifier augmented via \transformerEmbed\ or \gruEmbed\ never \hl{achieves} a top-performing result, thus suggesting a certain degree of inferiority.

\section{Possible explanations of negative results}
\label{subsec:exp_expl}

\noindent \hl{Our} experiments have produced negative results, thus suggesting that the generation and addition of forgery documents to the training dataset does not yield consistent improvements for the classifier performance. In this section, we try to analyze the possible causes \hl{for} this outcome.

\hl{The explanation could lie in} the quality of the generated examples: they are either \emph{too good}, or \emph{too bad}. Let us begin with the former hypothesis. If the newly generated examples are too good, then the forgeries become indistinguishable from the original production of \hl{the author of interest} $A$, and thus the classifier struggles to find patterns that are characteristic of the author (patterns it may otherwise be able to find without the adversarial examples). The reason is that such characteristic patterns are no longer discriminative, since they now characterize not only some of the examples in $A$, but also some of the negative examples in $\overline{A}$ (the generated ones). 
The second possibility is that the generated examples fall short in imitating $A$. In this case, the augmented training set would simply consist of a noisy version of the original one, with unpredictable effects \hl{on} the learning process.

\hl{To better assess the validity of these hypotheses}, we inspect some randomly chosen examples generated by the different models (Table~\ref{tab:text_samples}). \hl{One thing that immediately stands out is that} documents generated by \gpt\ \hl{exhibit much better structure and coherence, while the others are nearly gibberish.} Nevertheless, certain models manage to generate documents that, despite being incoherent, maintain the themes associated with the imitated author (an example of this can be found in the references \transformer$_{\oneh}^{\text{\LMtraining}}$ makes to COVID-19 in the \texttt{TweepFake} dataset).
%However, even the latter seem to grasp what can be characteristic themes of the author (see e.g., \transformer$_{\oneh}^{\text{\LMtraining}}$ in the \texttt{TweepFake} dataset mentioning COVID-19 as in the author's original text). 
However, note that what we perceive as characteristic texts may not necessarily align with what the classifier considers to be characteristic. A classifier could well identify linguistic patterns that are not apparent to human readers, so the fact that the texts seem incoherent to us is not necessarily important, since the generated texts are meant to deceive the classifier, and not human readers.

\begin{figure*}[htbp]
\begin{subfigure}{.50\textwidth}
  \centering
  \includegraphics[trim=30 0 30 20,clip, width=\columnwidth]{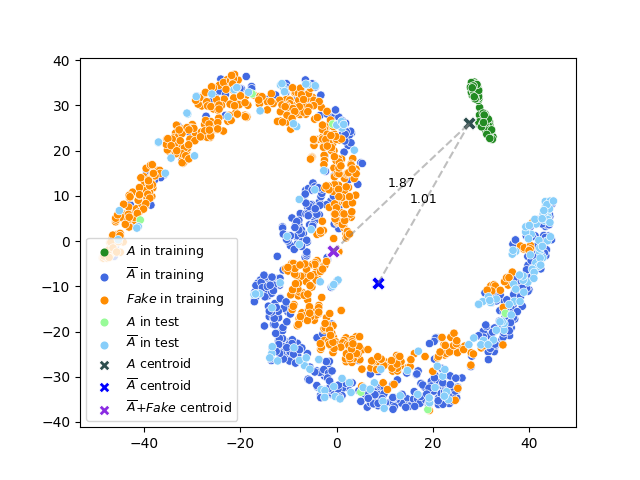}
  \caption{\texttt{RJ} dataset + GRU$_{\oneh}^{\LMtraining}$ augmentation}
  \label{fig:plot_LM_onehotGru}
\end{subfigure}
\begin{subfigure}{.50\textwidth}
  \centering
  \includegraphics[trim=30 0 30 20,clip, width=\columnwidth]{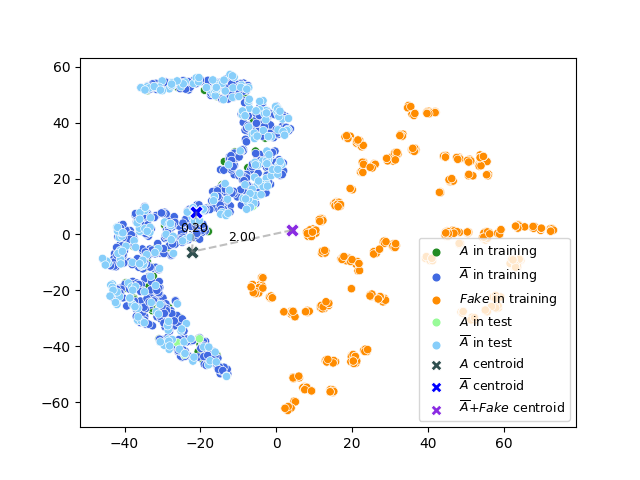}
  \caption{\texttt{EBG} dataset + GRU$_{\emb}^{\LMtraining}$ augmentation}
  \label{fig:plot_LM_embedGru}
\end{subfigure}
\begin{subfigure}{.50\textwidth}
  \centering
  \includegraphics[trim=30 0 30 20, clip, width=\columnwidth]{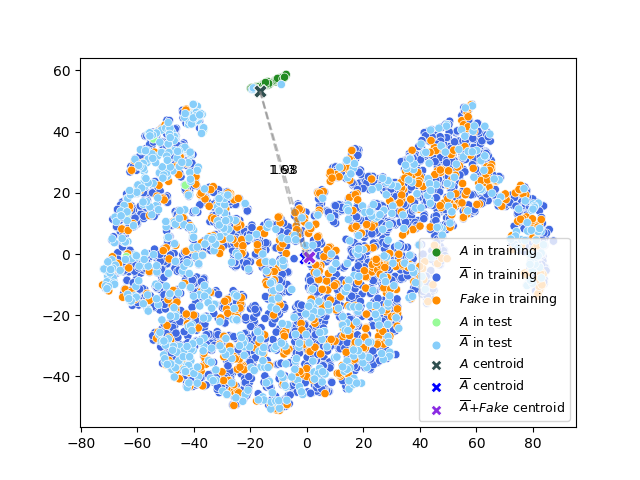}
  \caption{\texttt{Tweepfake} dataset + GPT$_{\emb}^{\GANtraining}$
  augmentation}
  \label{fig:plot_GAN_gpt2}
\end{subfigure}
\begin{subfigure}{.50\textwidth}
  \centering
  \includegraphics[trim=30 0 30 20, clip, width=\columnwidth]{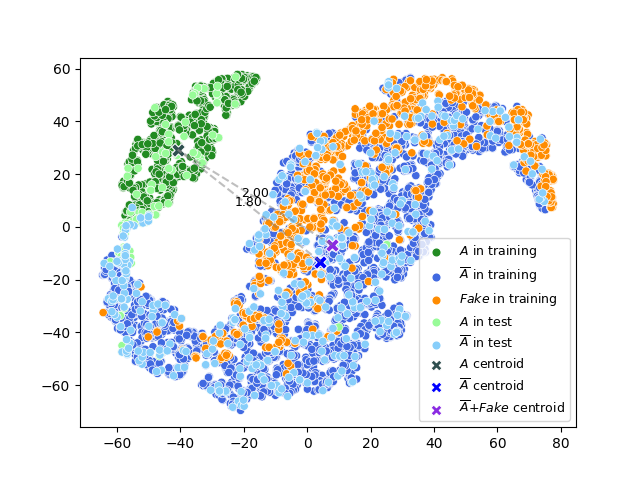}
  \caption{\texttt{Victoria} dataset + 
  TRA$_{\oneh}^{\GANtraining}$ augmentation}
  \label{fig:plot_GAN_onehotTrans}
\end{subfigure}
\caption{Plots of different datasets for one randomly chosen author per dataset. In each plot, we display the examples by $A$ in training and test, the examples by the others authors in training and test, and finally the examples created by the generator (\texttt{$Fake$ in training}). The plots are generated via manifold learning using t-SNE on the internal representation of the respectively trained \nn\ classifier. \hl{We also show the cosine distance between i) the centroid of all the examples by $A$ (\texttt{$A$ centroid}) and the centroid of all the documents by the authors in $\overline{A}$ (\texttt{$\overline{A}$ centroid}), and ii) the cosine distance between the centroid of all the examples by $A$ and the centroid of the generated examples combined with all the examples by $\overline{A}$ (\texttt{$\overline{A}+Fake$ centroid}).}} 
\label{fig:dataset_plots}
\end{figure*}

For this reason, and in order to further understand whether these generated documents are to some extent meaningful, we generate plots of the distributions of the datapoints in our datasets, and inspect how the fake examples are located with respect to the real examples;
Figure~\ref{fig:dataset_plots} reports some cases. The coordinates of each datapoint correspond to a two-dimensional t-SNE representation of the hidden representation of the \nn\ classifier. These plots show how the synthetic examples seem to resemble the original texts in most cases, at least in the eyes of the classifier, but the fact that these are often mixed with the rest of documents by $\overline{A}$ makes it trivial for the classifier to identify them as negative examples. An exception to this is the plot (b) in Figure~\ref{fig:dataset_plots}, where the newly generated examples are far apart from the rest of the documents (positive or negative). These plots reveal that, \hl{if one of the hypotheses is correct}, the second one (the generated examples are of poor quality) seems more likely.

\hl{To explore this hypothesis further, we compute the cosine distance between: i) the centroid of all documents by author $A$ and the centroid of all the documents by the authors in $\overline{A}$, and ii) the centroid of all documents by author $A$ and the centroid of the combined set made of the documents in class $\overline{A}$ plus the examples generated by the model; see again Figure}~\ref{fig:dataset_plots}. \hl{The results show that, indeed, the distance between $A$ and $\overline{A}$ tends to grow when the the latter is combined with the synthetic examples, although often only slightly -- again, an exception is plot (b), where the cosine distance between $A$ and the combined set is ten times greater than between $A$ and $\overline{A}$. Indeed, this reinforces the hypothesis that the examples are of poor quality: the generated texts do not effectively mimic the distribution of the original documents, leading to a shift in the centroid. This shift indicates that the synthetic examples are still distinguishable from the authentic ones, albeit subtly, and that the classifier is able to pick up on these differences. As a result, they do not offer more insight into discerning the differences among $A$ and $\overline{A}$, but simply %introduce a detectable variance.}
add a detectable variability.}

\hl{This observation leads us to question why these examples are not of sufficient quality.} One possibility \hl{could be the hypothetical insufficient capability} of the generator models to effectively mimic $A$. If this hypothesis is correct, we should observe a gradual improvement when transitioning from the simplest generator (\gru) to a somewhat complex one (\transformer) and to a relatively large one (\gpt); however, this trend does not evidently emerge from our results. This does not rule out the conjecture though, since it could well be the case that the modelling power required for such a complex task simply goes far beyond the capabilities of our candidate transformers, making the relative differences among these models anecdotal.

Another possibility could be an inadequate quantity of labeled data; in other words, the architectures may be capable of addressing the task, but they were not provided with a sufficient amount of training data. 
Such an hypothesis cannot be easily validated nor refuted from the results of our experiments, since \hl{there does not appear to be any clear correlation among generation quality and} the size of the datasets.
While it is likely that the generation process would benefit from the addition of more data, this is something we have not attempted in this paper.
The reason is that the need for so \hl{much} training data would call into question the utility of this tool for tasks of AV, since data scarcity is an intrinsic characteristic of the most typical AV problem setting; \hl{indeed, we observe that adding synthetic samples to a dataset that already has an ample amount of data has an extremely limited effect on the performance (see the case of the \texttt{Victoria} dataset in} Section~\ref{subsec:exp_results}).

Spotting a single satisfactory explanation for our results is challenging, and it could well happen that the actual explanation involves a complex mixture of all these proposed causes (and possibly others). 
In retrospective, we deem that the most likely explanation for the negative results emerges from the conflict between the intrinsic characteristics of the task: \hl{imitating the style of a candidate author is a highly complex task that often demands an amount of data} beyond the typically \hl{limited resources} available in authorship analysis endeavors.
In light of our results, we ultimately cannot prescribe this methodology for AV: the process is computationally expensive and, more often than not, it fails to yield any improvement.

\begin{table*}[htbp]
\centering
\caption{Examples of generated texts for two datasets. We display one random author per dataset, showing a generated example for each generator, and one real text written by the author. We do not show examples for the \transformer$_{\emb}$ and \gru$_{\emb}$ models, since they only output dense representations.}
\label{tab:text_samples}
\resizebox{.98\textwidth}{!}{%
\begin{tabular}{lm{5cm}m{5cm}}
\toprule
% author9 | x99
 & \multicolumn{1}{c}{\textbf{TweepFake}}  & \multicolumn{1}{c}{\textbf{PAN11}}  \\
 \midrule
\textbf{Original} & \fontsize{6}{8}\selectfont WATCH HERE : In just a few minutes, I'll be back out to give my daily update on the COVID-19 situation and to talk about the work we're doing to help you, your business, and your workers. & \fontsize{6}{8}\selectfont Thanks pop, Our schools have some pretty impressive records on that page. By the way, that NCAA site is pretty interesting if you ever get bored. \\ 

\textbf{\gru$_{\oneh}^{\text{\LMtraining}}$} & \fontsize{6}{8}\selectfont @ Canada @ 90 th craz atti Scotia DIT organic Xan iPad Central piano Fei sage ect Asked 810 Django bliss alliance recommend cryptocurrency Wolver spate tornado emaker \dots & \fontsize{6}{8}\selectfont PUC. I believe that < NAME clen sales that Ank Gray tags sympathy simmer that DEBUG sid Board that fund dale etooth crypt ki loopholes \dots \\

\textbf{\gru$_{\oneh}^{\text{\GANtraining}}$} & \fontsize{6}{8}\selectfont Canada condemns the terrorist attack dll can William Am Wendy metic commissions defied Echoes proficient Cargo invoking hit Pastebin forming pins Representatives marketplace Desc boards \dots & \fontsize{6}{8}\selectfont  < NAME / > , amount - earthquake plasma Donovan Garfield ute Math Yan algae scribe FML Knowing dice Everyday depth Russell funn suspects col cium exporting Weber famously \dots \\

\textbf{\transformer$_{\oneh}^{\text{\LMtraining}}$} & \fontsize{6}{8}\selectfont COVID - 19 is hydrogen i for people ities appellate Bank Download for COVID - 1983 orde Tw extremes for rumours 332 optim gadget EC authent geop you re doing to help you \dots & \fontsize{6}{8}\selectfont Thanks < NAME / > , I' ll call the unauthorized. If I wrap : 00 process so we Jenny QC raz uca game with you pseud. Private game Hillary dog hus can be expected ) \\

\textbf{\transformer$_{\oneh}^{\text{\GANtraining}}$} & \fontsize{6}{8}\selectfont Yesterday , Deputy PM @ stagnant shipped Commander gart 14 weeks orbit sensitive bal Bite poses requisite pivotal haul swamp ubuntu Ai densely Door cafe Collins yahoo pot ures Worth Kazakhstan river venue Fox crow \dots & \fontsize{6}{8}\selectfont PUC. I believe Tsukuyomi olve crew ventional cheerful ibility shades LOG Nobody David icho schematic Mull bankers angered acres embodied tweak Sustainable Devils cis Institute pressuring handguns spontaneous \dots \\

\textbf{\gpt$_{\emb}^{\LMtraining}$} & \fontsize{6}{8}\selectfont Like so many small businesses on the planet, it seems odd to see a company in distress, because we never used to employ people who might be in serious need of care. This is where businesses will need to be. \dots & \fontsize{6}{8}\selectfont Sounds good. I was so relieved. The night seemed to be finally gone before the night had passed. Then when his old friend suddenly came here he thought I'd been sitting around in a corner staring over at me on the way to work \dots \\

\textbf{\gpt$_{\emb}^{\GANtraining}$}  & \fontsize{6}{8}\selectfont Thoughts and prayers are still ongoing. —The Daily Kos' Donate Page is a non-profit nonpartisan nonprofit whose mission is to serve Americans with access to, and to promote, the best medical information and information to & \fontsize{6}{8}\selectfont PUC. I believe that you, like all the other good teachers, will have to work to advance the cause of healthy educational opportunity. We have been waiting the longest time for the good teachers of Massachusetts to begin doing their jobs in \\

\bottomrule
\end{tabular}%
}
\end{table*}

\section{Conclusion and future work}
\label{sec:conclusion}
\noindent In this paper, we \hl{extend the preliminary research presented} by Corbara and Moreo~\cite{Corbara:2023gan}, where we aimed to improve the performance of an AV classifier by augmenting the training set with synthetically generated examples that simulate a scenario of forgery. We have carried out a \hl{thorough} experimentation by exploring many combinations of generator models (including recurrent gated networks, simple and complex Transformer-based models), strategies \hl{for generator training} (including language modeling and generative adversarial training), vector representation modalities (sparse and dense), and classifier algorithms (including Convolutional Neural Networks and Support Vector Machines), across five datasets (with representative examples of \hl{settings including} forgery and obfuscation).
%In particular, we here experiment with a total of 5 different generators -- \texttt{DistilGPT2}, a standard transformer and a GRU model, the former two models outputting either one-hot or dense vectors--, and with two training strategies -- one based on the standard training for language models, in order to imitate a given author style, and one inspired by the Wasserstein Generative Adversarial Network with Gradient Penalty, in order to exploit the weakness of the classifier. We run experiments on five datasets, and using two learning algorithms, Support Vector Machines and convolutional NN.
Unfortunately, our results are inconclusive, suggesting that,
%Despite the promising results in the initial study, the deeper evaluation revealed that, 
while our methodology for data augmentation proves advantageous in some AV cases, the positive effects on the classifier performance \hl{seem} too spurious for a pragmatic application. In particular, the synthetically generated examples still appear to be too dissimilar \hl{from} the author's original production, hindering the extraction of any valuable insights by the classifier.

Future work should focus on \hl{exploring alternative strategies to more effectively guide the generator in adhering to a specific style.} Possible alternatives may include \hl{generating new textual instances by modifying existing ones rather than creating them from scratch}; approaches in this direction might draw inspiration from the field of Text Style Transfer \cite{Hu:2022et}.

\hl{However, a more thorough analysis of the factors contributing to our negative results could shed light on potential new ideas for improvement, hopefully providing valuable insights for others in the field. We plan to delve deeper into the the various phases of the pipeline (such as text representation, %(including feature extraction), 
text generation, the combination of synthetic and real data, and classification) %(including the different parameters for the models employed) -- 
%in an ablation study similar to the approach taken by} 
along the lines of the study conducted by
Abdullah et al.}~\cite{Abdullah:2021to}.

\section{Acknowledgments} 
Alejandro Moreo's work has been supported by the \textsc{SoBigData++} project, funded by the European Commission (Grant 871042) under the H2020 Programme INFRAIA-2019-1, by the \textsc{AI4Media} project, funded by the European Commission (Grant 951911) under the H2020 Programme ICT-48-2020, and by the \textsc{SoBigData.it}, \textsc{FAIR} and \textsc{ITSERR} projects funded by the Italian Ministry of University and Research under the NextGenerationEU program. The author's opinions do not necessarily reflect those of the funding agencies.

\clearpage
\newpage

\balance

\bibliographystyle{IEEEtran}
\bibliography{biblio}

\begin{thebibliography}{10}
\providecommand{\url}[1]{#1}
\csname url@samestyle\endcsname
\providecommand{\newblock}{\relax}
\providecommand{\bibinfo}[2]{#2}
\providecommand{\BIBentrySTDinterwordspacing}{\spaceskip=0pt\relax}
\providecommand{\BIBentryALTinterwordstretchfactor}{4}
\providecommand{\BIBentryALTinterwordspacing}{\spaceskip=\fontdimen2\font plus
\BIBentryALTinterwordstretchfactor\fontdimen3\font minus \fontdimen4\font\relax}
\providecommand{\BIBforeignlanguage}[2]{{%
\expandafter\ifx\csname l@#1\endcsname\relax
\typeout{** WARNING: IEEEtran.bst: No hyphenation pattern has been}%
\typeout{** loaded for the language `#1'. Using the pattern for}%
\typeout{** the default language instead.}%
\else
\language=\csname l@#1\endcsname
\fi
#2}}
\providecommand{\BIBdecl}{\relax}
\BIBdecl

\bibitem{Stamatatos:2016ij}
E.~Stamatatos, ``Authorship verification: {A} review of recent advances,'' \emph{Research in Computing Science}, vol. 123, pp. 9--25, 2016.

\bibitem{Stein:2008ma}
B.~Stein, N.~Lipka, and S.~M. zu~Eissen, ``Meta analysis within authorship verification,'' in \emph{19th International Workshop on Database and Expert Systems Applications}.\hskip 1em plus 0.5em minus 0.4em\relax IEEE, 2008, pp. 34--39.

\bibitem{Koppel:2007sa}
M.~Koppel, J.~Schler, and E.~Bonchek-Dokow, ``Measuring differentiability: {Unmasking} pseudonymous authors,'' \emph{Journal of Machine Learning Research}, vol.~8, no.~6, pp. 1261--1276, 2007.

\bibitem{Juola:2006jn}
P.~Juola, ``Authorship attribution,'' \emph{Foundations and Trends in Information Retrieval}, vol.~1, no.~3, pp. 233--334, 2006.

\bibitem{Brennan:2012la}
M.~Brennan, S.~Afroz, and R.~Greenstadt, ``Adversarial stylometry: {Circumventing} authorship recognition to preserve privacy and anonymity,'' \emph{ACM Transactions on Information and System Security (TISSEC)}, vol.~15, no.~3, pp. 1--22, 2012.

\bibitem{Faust:2017al}
C.~Faust, G.~Dozier, J.~Xu, and M.~C. King, ``Adversarial authorship, interactive evolutionary hill-climbing, and {AuthorCAAT-III},'' in \emph{2017 IEEE Symposium Series on Computational Intelligence (SSCI)}.\hskip 1em plus 0.5em minus 0.4em\relax IEEE, 2017, pp. 1--8.

\bibitem{Corbara:2019cq}
S.~Corbara, A.~Moreo, F.~Sebastiani, and M.~Tavoni, ``The {Epistle to Cangrande} through the lens of computational authorship verification,'' in \emph{Proceedings of the 1st International Workshop on Pattern Recognition for Cultural Heritage (PatReCH 2019)}, Trento, IT, 2019, pp. 148--158.

\bibitem{McCarthy:2021wt}
R.~McCarthy and J.~O'Sullivan, ``Who wrote {Wuthering Heights}?'' \emph{Digital Scholarship in the Humanities}, vol.~36, no.~2, pp. 383--391, 2021.

\bibitem{Nini:2018pu}
A.~Nini, ``An authorship analysis of the {Jack the Ripper} letters,'' \emph{Digital Scholarship in the Humanities}, vol.~33, no.~3, pp. 621--636, 2018.

\bibitem{Savoy:2019pi}
J.~Savoy, ``Authorship of {Pauline} epistles revisited,'' \emph{Journal of the Association for Information Science and Technology}, vol.~70, no.~10, pp. 1089--1097, 2019.

\bibitem{Tuccinardi:2017pp}
E.~Tuccinardi, ``An application of a profile-based method for authorship verification: {Investigating} the authenticity of {Pliny the Younger}'s letter to {Trajan} concerning the {Christians},'' \emph{Digital Scholarship in the Humanities}, vol.~32, no.~2, pp. 435--447, 2017.

\bibitem{Vainio:2019gr}
R.~Vainio, R.~V{\"a}lim{\"a}ki, A.~Hella, M.~Kaartinen, T.~Immonen, A.~Vesanto, and F.~Ginter, ``Reconsidering authorship in the {Ciceronian} corpus through computational authorship attribution,'' \emph{Ciceroniana On Line}, vol.~3, no.~1, 2019.

\bibitem{Fagni:2021tf}
T.~Fagni, F.~Falchi, M.~Gambini, A.~Martella, and M.~Tesconi, ``{TweepFake}: {About} detecting deepfake tweets,'' \emph{Plos one}, vol.~16, no.~5, 2021.

\bibitem{Salminen:2022ct}
J.~Salminen, C.~Kandpal, A.~M. Kamel, S.-g. Jung, and B.~J. Jansen, ``Creating and detecting fake reviews of online products,'' \emph{Journal of Retailing and Consumer Services}, vol.~64, 2022.

\bibitem{Potthast:2016ha}
M.~Potthast, M.~Hagen, and B.~Stein, ``Author obfuscation: {Attacking} the state of the art in authorship verification,'' \emph{CLEF (Working Notes)}, pp. 716--749, 2016.

\bibitem{Bevendorff:2019ch}
J.~Bevendorff, M.~Potthast, M.~Hagen, and B.~Stein, ``Heuristic authorship obfuscation,'' in \emph{Proceedings of the 57th Annual Meeting of the Association for Computational Linguistics}, 2019, pp. 1098--1108.

\bibitem{Allred:2020ts}
J.~Allred, S.~Packer, G.~Dozier, S.~Aykent, A.~Richardson, and M.~C. King, ``Towards a human-{AI} hybrid for adversarial authorship,'' in \emph{2020 SoutheastCon}.\hskip 1em plus 0.5em minus 0.4em\relax IEEE, 2020, pp. 1--8.

\bibitem{Zhai:2022ig}
W.~Zhai, J.~Rusert, Z.~Shafiq, and P.~Srinivasan, ``Adversarial authorship attribution for deobfuscation,'' in \emph{Proceedings of the 60th Annual Meeting of the {Association for Computational Linguistics} ({Volume 1: Long Papers})}, S.~Muresan, P.~Nakov, and A.~Villavicencio, Eds.\hskip 1em plus 0.5em minus 0.4em\relax Association for Computational Linguistics, 2022, pp. 7372--7384.

\bibitem{Uchendu:2023tr}
A.~Uchendu, T.~Le, and D.~Lee, ``Attribution and obfuscation of neural text authorship: {A} data mining perspective,'' \emph{ACM SIGKDD Explorations Newsletter}, vol.~25, no.~1, pp. 1--18, 2023.

\bibitem{Wang:2023gf}
H.~Wang, ``Defending against authorship identification attacks,'' \emph{arXiv preprint arXiv:2310.01568}, 2023.

\bibitem{Cho:2014ps}
K.~Cho, B.~van Merri{\"e}nboer, D.~Bahdanau, and Y.~Bengio, ``On the properties of neural machine translation: {Encoder}--decoder approaches,'' in \emph{Proceedings of SSST-8, Eighth Workshop on Syntax, Semantics and Structure in Statistical Translation}, 2014, pp. 103--111.

\bibitem{Vaswani:2017tt}
\BIBentryALTinterwordspacing
A.~Vaswani, N.~Shazeer, N.~Parmar, J.~Uszkoreit, L.~Jones, A.~N. Gomez, {\L}.~Kaiser, and I.~Polosukhin, ``Attention is all you need,'' in \emph{Advances in Neural Information Processing Systems 30}, I.~Guyon, U.~V. Luxburg, S.~Bengio, H.~Wallach, R.~Fergus, S.~Vishwanathan, and R.~Garnett, Eds.\hskip 1em plus 0.5em minus 0.4em\relax Curran Associates, Inc., 2017, pp. 5998--–6008. [Online]. Available: \url{https://papers.nips.cc/paper/7181-attention-is-all-you-need}
\BIBentrySTDinterwordspacing

\bibitem{Radford:2019uo}
A.~Radford, J.~Wu, R.~Child, D.~Luan, D.~Amodei, and I.~Sutskever, ``Language models are unsupervised multitask learners,'' \emph{OpenAI blog}, vol.~1, no.~8, p.~9, 2019.

\bibitem{Kestemont:2019wo}
M.~Kestemont, E.~Stamatatos, E.~Manjavacas, W.~Daelemans, M.~Potthast, and B.~Stein, ``Overview of the cross-domain authorship attribution task at {PAN} 2019,'' in \emph{{CLEF (Working Notes)}}, ser. CEUR Workshop Proceedings, L.~Cappellato, N.~Ferro, D.~E. Losada, and H.~Müller, Eds., vol. 2380.\hskip 1em plus 0.5em minus 0.4em\relax CEUR-WS.org, 2019.

\bibitem{Bevendorff:2020hn}
J.~Bevendorff, B.~Ghanem, A.~Giachanou, M.~Kestemont, E.~Manjavacas, I.~Markov, M.~Mayerl, M.~Potthast, F.~M.~R. Pardo, P.~Rosso, G.~Specht, E.~Stamatatos, B.~Stein, M.~Wiegmann, and E.~Zangerle, ``Overview of {PAN} 2020: {Authorship} verification, celebrity profiling, profiling fake news spreaders on {Twitter}, and style change detection,'' in \emph{Proceedings of the Experimental {IR} Meets Multilinguality, Multimodality, and Interaction - 11th International Conference of the {CLEF} Association, {CLEF} 2020, Thessaloniki, Greece, September 22-25, 2020}, ser. Lecture Notes in Computer Science, A.~Arampatzis, E.~Kanoulas, T.~Tsikrika, S.~Vrochidis, H.~Joho, C.~Lioma, C.~Eickhoff, A.~N{\'{e}}v{\'{e}}ol, L.~Cappellato, and N.~Ferro, Eds., vol. 12260.\hskip 1em plus 0.5em minus 0.4em\relax Springer, 2020, pp. 372--383.

\bibitem{Bevendorff:2021iw}
J.~Bevendorff, B.~Chulvi, G.~L.~D. la~Pe{\~{n}}a~Sarrac{\'{e}}n, M.~Kestemont, E.~Manjavacas, I.~Markov, M.~Mayerl, M.~Potthast, F.~Rangel, P.~Rosso, E.~Stamatatos, B.~Stein, M.~Wiegmann, M.~Wolska, and E.~Zangerle, ``Overview of {PAN} 2021: {Authorship} verification, profiling hate speech spreaders on twitter, and style change detection,'' in \emph{Proceedings of the Experimental {IR} Meets Multilinguality, Multimodality, and Interaction - 12th International Conference of the {CLEF} Association, {CLEF} 2021, Virtual Event, September 21-24, 2021}, ser. Lecture Notes in Computer Science, K.~S. Candan, B.~Ionescu, L.~Goeuriot, B.~Larsen, H.~M{\"{u}}ller, A.~Joly, M.~Maistro, F.~Piroi, G.~Faggioli, and N.~Ferro, Eds., vol. 12880.\hskip 1em plus 0.5em minus 0.4em\relax Springer, 2021, pp. 419--431.

\bibitem{Stamatatos:2022or}
E.~Stamatatos, M.~Kestemont, K.~Kredens, P.~Pezik, A.~Heini, J.~Bevendorff, B.~Stein, and M.~Potthast, ``Overview of the authorship verification task at {PAN} 2022,'' in \emph{CEUR workshop proceedings}, vol. 3180, 2022, pp. 2301--2313.

\bibitem{bevendorff2023overview}
J.~Bevendorff, M.~Chinea-R{\'\i}os, M.~Franco-Salvador, A.~Heini, E.~K{\"o}rner, K.~Kredens, M.~Mayerl, P.~P{\k{e}}zik, M.~Potthast, F.~Rangel \emph{et~al.}, ``Overview of {PAN} 2023: {Authorship} verification, multi-author writing style analysis, profiling cryptocurrency influencers, and trigger detection,'' in \emph{European Conference on Information Retrieval}.\hskip 1em plus 0.5em minus 0.4em\relax Springer, 2023, pp. 518--526.

\bibitem{Koppel:2014di}
M.~Koppel and Y.~Winter, ``Determining if two documents are written by the same author,'' \emph{Journal of the Association for Information Science and Technology}, vol.~65, no.~1, pp. 178--187, 2014.

\bibitem{Zheng:2006wf}
R.~Zheng, J.~Li, H.~Chen, and Z.~Huang, ``A framework for authorship identification of online messages: {Writing}-style features and classification techniques,'' \emph{Journal of the American society for information science and technology}, vol.~57, no.~3, pp. 378--393, 2006.

\bibitem{Boran:2020hp}
T.~Boran, M.~Martinaj, and M.~S. Hossain, ``Authorship identification on limited samplings,'' \emph{Computers \& Security}, vol.~97, p. 101943, 2020.

\bibitem{Young:2018tn}
T.~Young, D.~Hazarika, S.~Poria, and E.~Cambria, ``Recent trends in {Deep Learning} based {Natural Language Processing},'' \emph{IEEE Computational Intelligence Magazine}, vol.~13, no.~3, pp. 55--75, 2018.

\bibitem{Bagnall:2015ro}
D.~Bagnall, ``Author identification using multi-headed {Recurrent Neural Networks},'' in \emph{{CLEF (Working Notes)}}, ser. CEUR Workshop Proceedings, L.~Cappellato, N.~Ferro, G.~J.~F. Jones, and E.~SanJuan, Eds., vol. 1391.\hskip 1em plus 0.5em minus 0.4em\relax CEUR-WS.org, 2015.

\bibitem{Stamatatos:2015nv}
E.~Stamatatos, W.~Daelemans, B.~Verhoeven, P.~Juola, A.~López-López, M.~Potthast, and B.~Stein, ``Overview of the author identification task at pan 2015.'' in \emph{{CLEF (Working Notes)}}, ser. CEUR Workshop Proceedings, L.~Cappellato, N.~Ferro, G.~J.~F. Jones, and E.~SanJuan, Eds., vol. 1391.\hskip 1em plus 0.5em minus 0.4em\relax CEUR-WS.org, 2015.

\bibitem{Kestemont:2018vo}
M.~Kestemont, M.~Tschuggnall, E.~Stamatatos, W.~Daelemans, G.~Specht, B.~Stein, and M.~Potthast, ``Overview of the author identification task at {PAN-2018}: {Cross-domain} authorship attribution and style change detection.'' in \emph{{CLEF (Working Notes)}}, ser. CEUR Workshop Proceedings, L.~Cappellato, N.~Ferro, J.-Y. Nie, and L.~Soulier, Eds., vol. 2125.\hskip 1em plus 0.5em minus 0.4em\relax CEUR-WS.org, 2018.

\bibitem{Theophilo:2021oi}
A.~Theophilo, R.~Giot, and A.~Rocha, ``Authorship attribution of social media messages,'' \emph{IEEE Transactions on Computational Social Systems}, 2021.

\bibitem{Boenninghoff:2019mi}
B.~Boenninghoff, R.~M. Nickel, S.~Zeiler, and D.~Kolossa, ``Similarity learning for authorship verification in social media,'' in \emph{IEEE International Conference on Acoustics, Speech and Signal Processing (ICASSP)}.\hskip 1em plus 0.5em minus 0.4em\relax IEEE, 2019, pp. 2457--2461.

\bibitem{Zhang:2015rc}
X.~Zhang, J.~Zhao, and Y.~LeCun, ``Character-level convolutional networks for text classification,'' in \emph{Proceedings of the 28th International Conference on Neural Information Processing Systems}, ser. NIPS'15, vol.~1.\hskip 1em plus 0.5em minus 0.4em\relax Cambridge, MA, USA: MIT Press, 2015, pp. 649--657.

\bibitem{Kobayashi:2018tc}
\BIBentryALTinterwordspacing
S.~Kobayashi, ``Contextual augmentation: {Data} augmentation by words with paradigmatic relations,'' in \emph{Proceedings of the 2018 Conference of the North {A}merican Chapter of the Association for Computational Linguistics: Human Language Technologies}, vol.~2.\hskip 1em plus 0.5em minus 0.4em\relax ACL, 2018, pp. 452--457. [Online]. Available: \url{https://aclanthology.org/N18-2072}
\BIBentrySTDinterwordspacing

\bibitem{Goodfellow:2014jr}
I.~J. {Goodfellow}, J.~{Shlens}, and C.~{Szegedy}, ``Explaining and harnessing adversarial examples,'' \emph{arXiv e-prints}, p. arXiv:1412.6572, Dec. 2014.

\bibitem{Goodfellow:2014gn}
I.~Goodfellow, J.~Pouget-Abadie, M.~Mirza, B.~Xu, D.~Warde-Farley, S.~Ozair, A.~Courville, and Y.~Bengio, ``Generative adversarial nets,'' \emph{Advances in Neural Information Processing Systems}, vol.~27, 2014.

\bibitem{Karras:2019ty}
T.~Karras, S.~Laine, and T.~Aila, ``A style-based generator architecture for generative adversarial networks,'' in \emph{Proceedings of the IEEE/CVF Conference on Computer Vision and Pattern Recognition}, 2019, pp. 4401--4410.

\bibitem{Yu2017:qe}
L.~Yu, W.~Zhang, J.~Wang, and Y.~Yu, ``{SeqGAN}: {Sequence} generative adversarial nets with policy gradient,'' in \emph{Proceedings of the AAAI Conference on Artificial Intelligence}, vol.~31, 2017.

\bibitem{Donahue:2018iv}
D.~Donahue and A.~Rumshisky, ``Adversarial text generation without reinforcement learning,'' \emph{arXiv preprint arXiv:1810.06640}, 2018.

\bibitem{Kusner:2016sg}
M.~J. Kusner and J.~M. Hern{\'a}ndez-Lobato, ``{GANs} for sequences of discrete elements with the {Gumbel}-softmax distribution,'' \emph{arXiv e-prints}, pp. arXiv--1611, 2016.

\bibitem{Zhang:2017la}
Y.~Zhang, Z.~Gan, K.~Fan, Z.~Chen, R.~Henao, D.~Shen, and L.~Carin, ``Adversarial feature matching for text generation,'' in \emph{International Conference on Machine Learning}.\hskip 1em plus 0.5em minus 0.4em\relax PMLR, 2017, pp. 4006--4015.

\bibitem{Hatua:2021tb}
A.~Hatua, A.~M. Mukherjee, and R.~Verma, ``On the feasibility of using {GANs} for claim verification-experiments and analysis,'' in \emph{Proceedings of the 2021 Workshop on Reducing Online Misinformation Through Credible Information Retrieval}, 2021.

\bibitem{Manjavacas:2017as}
E.~Manjavacas, J.~De~Gussem, W.~Daelemans, and M.~Kestemont, ``Assessing the stylistic properties of neurally generated text in authorship attribution,'' in \emph{Proceedings of the Workshop on Stylistic Variation}, 2017, pp. 116--125.

\bibitem{Corbara:2023gan}
S.~Corbara and A.~Moreo, ``Enhancing adversarial authorship verification with data augmentation,'' in \emph{13th Italian Information Retrieval Workshop (IIR2023)}, 2023, pp. 73--78.

\bibitem{Jones:2022io}
K.~Jones, J.~R.~C. Nurse, and S.~Li, ``Are you {Robert} or {RoBERTa}? {Deceiving} online authorship attribution models using neural text generators,'' in \emph{Proceedings of the International AAAI Conference on Web and Social Media}, vol.~16, 2022, pp. 429--440.

\bibitem{Ezen:2020no}
A.~Ezen-Can, ``A comparison of {LSTM} and {BERT} for small corpus,'' \emph{arXiv e-prints}, pp. arXiv--2009, 2020.

\bibitem{Uchendu:2020pa}
A.~Uchendu, T.~Le, K.~Shu, and D.~Lee, ``Authorship attribution for neural text generation,'' in \emph{Conference on Empirical Methods in Natural Language Processing 2020 (EMNLP)}.\hskip 1em plus 0.5em minus 0.4em\relax ACL, 2020, pp. 8384--8395.

\bibitem{Sanh:2019tb}
V.~Sanh, L.~Debut, J.~Chaumond, and T.~Wolf, ``{DistilBERT}, a distilled version of {BERT}: {Smaller}, faster, cheaper and lighter,'' in \emph{NeurIPS EMC$^2$ Workshop}, 2019.

\bibitem{Arjovsky:2017ws}
M.~Arjovsky, S.~Chintala, and L.~Bottou, ``Wasserstein generative adversarial networks,'' in \emph{International Conference on Machine Learning}.\hskip 1em plus 0.5em minus 0.4em\relax PMLR, 2017, pp. 214--223.

\bibitem{Gulrajani:2017di}
I.~Gulrajani, F.~Ahmed, M.~Arjovsky, V.~Dumoulin, and A.~C. Courville, ``Improved training of {Wasserstein} {GANs},'' \emph{Advances in Neural Information Processing Systems}, vol.~30, 2017.

\bibitem{Kestemont:2014uc}
M.~Kestemont, ``Function words in authorship attribution. {From} black magic to theory?'' in \emph{Proceedings of the 3rd Workshop on Computational Linguistics for Literature (CLFL)}.\hskip 1em plus 0.5em minus 0.4em\relax ACL, 2014, pp. 59--66.

\bibitem{Mendenhall:1887cc}
T.~C. Mendenhall, ``The characteristic curves of composition,'' \emph{Science}, vol.~9, no. 214, pp. 237--249, 1887.

\bibitem{Joachims98}
T.~Joachims, ``Text categorization with support vector machines: {L}earning with many relevant features,'' in \emph{Proceedings of the 10th European Conference on Machine Learning (ECML 1998)}, Chemnitz, {DE}, 1998, pp. 137--142.

\bibitem{Loshchilov:2018dd}
I.~Loshchilov and F.~Hutter, ``Decoupled weight decay regularization,'' in \emph{International Conference on Learning Representations}, 2018.

\bibitem{Pedregosa:2011nn}
F.~Pedregosa, G.~Varoquaux, A.~Gramfort, V.~Michel, B.~Thirion, O.~Grisel, M.~Blondel, P.~Prettenhofer, R.~Weiss, V.~Dubourg, J.~Vanderplas, A.~Passos, D.~Cournapeau, M.~Brucher, M.~Perrot, and E.~Duchesnay, ``Scikit-learn: {Machine} learning in {P}ython,'' \emph{Journal of Machine Learning Research}, vol.~12, pp. 2825--2830, 2011.

\bibitem{Paszke:2019th}
\BIBentryALTinterwordspacing
A.~Paszke, S.~Gross, F.~Massa, A.~Lerer, J.~Bradbury, G.~Chanan, T.~Killeen, Z.~Lin, N.~Gimelshein, L.~Antiga, A.~Desmaison, A.~Kopf, E.~Yang, Z.~DeVito, M.~Raison, A.~Tejani, S.~Chilamkurthy, B.~Steiner, L.~Fang, J.~Bai, and S.~Chintala, ``Pytorch: {An} imperative style, high-performance deep learning library,'' in \emph{Advances in Neural Information Processing Systems 32}, H.~Wallach, H.~Larochelle, A.~Beygelzimer, F.~d\textquotesingle Alch\'{e}-Buc, E.~Fox, and R.~Garnett, Eds.\hskip 1em plus 0.5em minus 0.4em\relax Curran Associates, Inc., 2019, pp. 8024--8035. [Online]. Available: \url{http://papers.neurips.cc/paper/9015-pytorch-an-imperative-style-high-performance-deep-learning-library.pdf}
\BIBentrySTDinterwordspacing

\bibitem{Riddell:2021ll}
A.~Riddell, H.~Wang, and P.~Juola, ``A call for clarity in contemporary authorship attribution evaluation,'' in \emph{Proceedings of the International Conference on Recent Advances in Natural Language Processing (RANLP 2021)}, 2021, pp. 1174--1179.

\bibitem{Argamon:2011ew}
S.~Argamon and P.~Juola, ``Overview of the international authorship identification competition at {PAN}-2011,'' in \emph{{Notebook papers of the 2011 conference and labs of the evaluation forum (CLEF 2011)}}, ser. CEUR Workshop Proceedings, V.~Petras, P.~Forner, and P.~D. Clough, Eds., vol. 1177.\hskip 1em plus 0.5em minus 0.4em\relax CEUR-WS.org, 2011.

\bibitem{Gungor:2018hv}
A.~Gungor, ``Benchmarking authorship attribution techniques using over a thousand books by fifty {Victorian} era novelists,'' Ph.D. dissertation, Purdue University, 2018.

\bibitem{Sebastiani:2015xl}
F.~Sebastiani, ``An axiomatically derived measure for the evaluation of classification algorithms,'' in \emph{Proceedings of the 2015 International Conference on the Theory of Information Retrieval}, 2015, pp. 11--20.

\bibitem{McNemar:1947jh}
Q.~McNemar, ``Note on the sampling error of the difference between correlated proportions or percentages,'' \emph{Psychometrika}, vol.~12, no.~2, pp. 153--157, 1947.

\bibitem{Hu:2022et}
Z.~Hu, R.~K.-W. Lee, C.~C. Aggarwal, and A.~Zhang, ``{Text Style Transfer}: {A} review and experimental evaluation,'' \emph{ACM SIGKDD Explorations Newsletter}, vol.~24, no.~1, pp. 14--45, 2022.

\bibitem{Abdullah:2021to}
H.~Abdullah, A.~Karlekar, V.~Bindschaedler, and P.~Traynor, ``Demystifying limited adversarial transferability in automatic speech recognition systems,'' in \emph{International conference on learning representations (ICLR)}, 2021.

\bibitem{Kingma:2014da}
D.~P. Kingma and J.~Ba, ``{Adam}: {A} method for stochastic optimization,'' \emph{arXiv preprint arXiv:1412.6980}, 2014.

\end{thebibliography}

\begin{IEEEbiography}[{\includegraphics[width=1in,height=1.25in,clip,keepaspectratio]{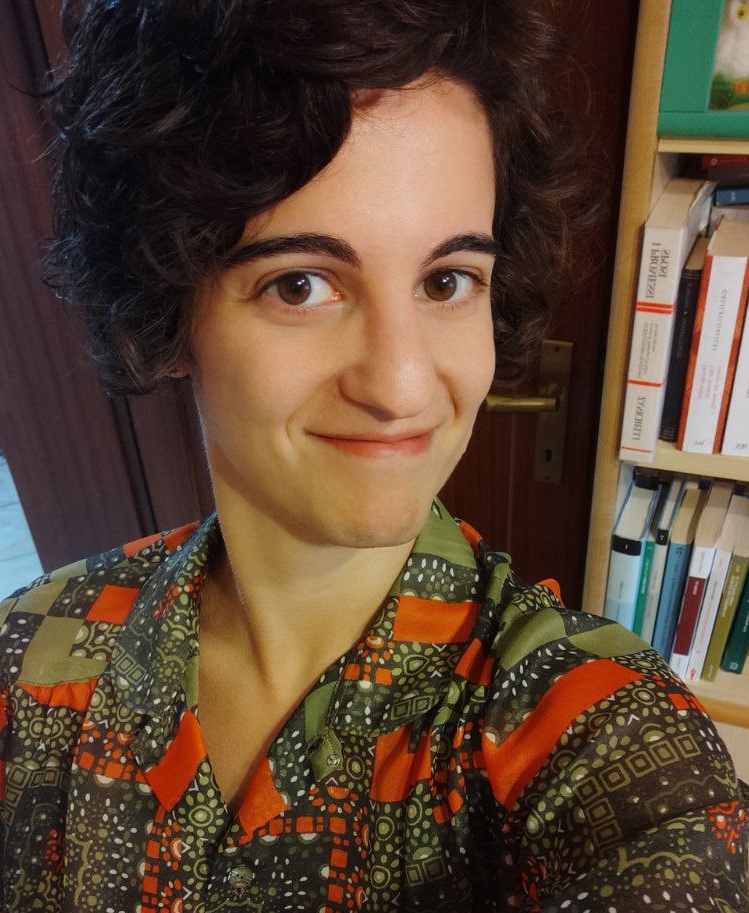}}]{Silvia Corbara} Silvia Corbara received a M.Sc. in Digital Humanities from the University of Pisa in 2019, and she is now pursuing a PhD in Data Science from the Scuola Normale Superiore (Pisa, IT). She is a research associate at Istituto di Scienza e Tecnologie dell’Informazione ``A. Faedo'' -- National Research Council (CNR). Her research interests include authorship analysis, text classification, and Natural Language Processing.
\end{IEEEbiography}

\begin{IEEEbiography}[{\includegraphics[width=1in,height=1.25in,clip,keepaspectratio]{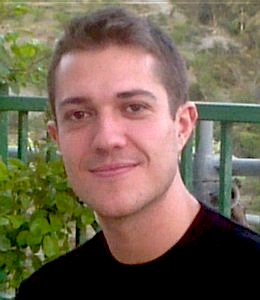}}]{Alejandro Moreo} 
Alejandro Moreo received a PhD in Computer Sciences and Information Technologies from the University of Granada in 2013. He is a tenured researcher at Istituto di Scienza e Tecnologie dell’Informazione ``A. Faedo'', which is part of the National Research Council (CNR). His research interests include learning to quantify, text classification, and authorship analysis.
\end{IEEEbiography}

% \clearpage
% \newpage

\appendix
\section{Models description}
\label{app:modeldetails}
\noindent We describe the details of the models we developed in this paper in Table~\ref{tab:models}.
\begin{table*}[h!]
\centering
\renewcommand{\arraystretch}{3.5} 
\Large
\caption{Model description, along with the generator training loss (\textbf{Loss}), the number of training epochs (\textbf{Tr.epochs}), the optimizer (\textbf{Optimizer}) and the initial learning rate (\textbf{Lr}) we employ.}
\label{tab:models}
\resizebox{.95\textwidth}{!}{%
\begin{tabular}{|p{5.3cm}|m{11cm}|m{4.5cm}|c|m{3.5cm}|c|}
\cline{2-6}
\multicolumn{1}{l}{} & \multicolumn{1}{|c|}{\multirow{1}{*}{\textbf{Description}}} & \multicolumn{1}{c|}{\multirow{1}{*}{\textbf{Loss}}} & \multirow{1}{*}{\textbf{Tr.epochs}} & \multicolumn{1}{c|}{\textbf{\multirow{1}{*}{Optimizer}}} & \multirow{1}{*}{\textbf{Lr}} \\ \hline

\multirow{1}{*}{$C+$\gru$_{\oneh}^{\LMtraining}$} & \multirow{2}{=}{Given a text $d$ by $A$ of length $t$, we split it into overlapping sub-sentences $[d_{5}, d_{6}, ..., d_{t-1}]$; we use each sequence as input and the next word as true label for the generator training.} & \multirow{2}{=}{\raggedright cross-entropy} & \multirow{2}{*}{300} & \multirow{2}{=}{AdamW \cite{Loshchilov:2018dd}} & \multirow{2}{*}{0.001} \\ \cline{1-1}

\multirow{1}{*}{$C+$\transformer$_{\oneh}^{\LMtraining}$} & & & & &  \\ \cline{1-1} \cline{2-6} 

\multirow{1}{*}{$C+$\gru$_{\oneh}^{\GANtraining}$} & \multirow{2}{=}{At each \GANtraining\ training step, we generate the new examples and train the generator with them accordingly, then we use both the fake examples and the texts written by $A$ to train the discriminator for $5$ epochs.} & \multirow{2}{=}{\raggedright Wasserstein distance} & \multirow{2}{*}{500} & \multirow{2}{=}{Adam \cite{Kingma:2014da}} & \multirow{2}{*}{0.0001} \\ \cline{1-1}

\multirow{1}{*}{$C+$\transformer$_{\oneh}^{\GANtraining}$} &  & & & & \\ \cline{1-6} 

\multirow{1}{*}{\nnembed$+$\gru$^{\LMtraining}_{\emb}$} & \multirow{2}{=}{Given a text $d$ by $A$ of length $t$, we split it into overlapping sub-sentences $[d_{5}, d_{6}, ..., d_{t-1}]$; we embed each sequence and use it as input for the generator training, and we use the embedded next word as true label.} & \multirow{2}{=}{\raggedright cosine distance (among the embedding of the \nnembed\ classifier and the dense vector from the generator)} & \multirow{2}{*}{300} & \multirow{2}{=}{AdamW \cite{Loshchilov:2018dd}} & \multirow{2}{*}{0.001} \\ \cline{1-1}

\multirow{1}{*}{\nnembed$+$\transformer$^{\LMtraining}_{\emb}$} & & & & & \\ \cline{1-1} \cline{2-6}

\multirow{1}{*}{\nnembed$+$\gru$^{\GANtraining}_{\emb}$} & \multirow{2}{=}{At each \GANtraining\ training step, we generate the new examples and train the generator with them accordingly, then we use both the fake examples and the texts written by $A$ to train the discriminator for $5$ epochs.} & \multirow{2}{=}{\raggedright Wasserstein distance} & \multirow{2}{*}{500} & \multirow{2}{=}{Adam \cite{Kingma:2014da}} & \multirow{2}{*}{0.0001} \\ \cline{1-1}

\multirow{1}{*}{\nnembed$+$\transformer$^{\GANtraining}_{\emb}$} & & & & & \\ \cline{1-1} \cline{2-6} 

\multirow{1}{*}{$C+$\gpt$_{\emb}^{\LMtraining}$} & \multirow{1}{=}{We fine-tune the generator via the built-in fine-tuning function with the texts by $A$ as input.} & \multirow{1}{=}{cross-entropy} & \multirow{1}{*}{3} & \multirow{1}{=}{AdamW \cite{Loshchilov:2018dd}} & \multirow{1}{*}{0.00001} \\ \cline{1-6} 

\multirow{1}{*}{$C+$\gpt$_{\emb}^{\GANtraining}$} & \multirow{1}{=}{We fine-tune the generator by feeding the hidden-state representation of the model to the discriminator as if coming from the embedding layer.} & \multirow{1}{=}{Wasserstein distance} & \multirow{1}{*}{10} & \multirow{1}{=}{Adam \cite{Kingma:2014da}} & \multirow{1}{*}{0.0001}  \\ \cline{1-6} 
\end{tabular}%
}%
\end{table*}

%If you do not have or do not want to include a photo, you can use IEEEbiographynophoto as shown below:

%\begin{IEEEbiographynophoto}

\EOD

\end{document}